# Generating Reliable Synthetic Clinical Trial Data: The Role of Hyperparameter Optimization and Domain Constraints

Waldemar Hahn[a,b]*, Jan-Niklas Eckardt[c,d], Christoph Röllig[c], Martin Sedlmayr[b], Jan Moritz Middeke[c,d], Markus Wolfien[b,a]

[a] Center for Scalable Data Analytics and Artificial Intelligence (ScaDS.AI) Dresden/Leipzig, Dresden, Germany
[b] Institute for Medical Informatics and Biometry, Technical University Dresden, Dresden, Germany
[c] Department of Internal Medicine I, University Hospital Carl Gustav Carus, Technical University Dresden, Dresden, Germany
[d] Else Kröner Fresenius Center for Digital Health, Technical University Dresden, Dresden, Germany

* Corresponding author: Waldemar Hahn, waldemar.hahn@tu-dresden.de

## Abstract

The generation of synthetic clinical trial data offers a promising approach to mitigating privacy concerns and data accessibility limitations in medical research. However, ensuring that synthetic datasets maintain high fidelity, utility, and adherence to domain-specific constraints remains a key challenge. While hyperparameter optimization (HPO) improves generative model performance, the effectiveness of different optimization strategies for synthetic clinical data remains unclear. This study systematically evaluates four HPO objectives across nine generative models, comparing single-metric to compound metric optimization. Our results demonstrate that HPO consistently improves synthetic data quality, with Tab DDPM achieving the largest relative gains, followed by TVAE (60%), CTGAN (39%), and CTAB-GAN+ (38%). Compound metric optimization outperformed single-metric objectives, producing more generalizable synthetic datasets. Despite improving overall quality, HPO alone fails to prevent violations of essential clinical survival constraints. Preprocessing and postprocessing played a crucial role in reducing these violations, as models lacking robust processing steps produced invalid data in up to 61% of cases. These findings underscore the necessity of integrating explicit domain knowledge alongside HPO to generate high-quality synthetic datasets. Our study provides actionable recommendations for improving synthetic data generation, with future work needed to refine metric selection and validate findings on larger datasets.

## 1. Introduction

Synthetic data generation has rapidly gained attention across various fields as a promising strategy to address data scarcity, privacy concerns, and restricted access [1], [2], [3]. In healthcare, particularly in clinical trials, regulatory and proprietary constraints often limit the sharing of patient-level information, complicating collaborative efforts. At the same time, high-quality datasets are essential not only for advancing clinical research but also for driving the development and evaluation of new algorithms. By mimicking the statistical and structural properties of real-world data while safeguarding sensitive information, synthetic datasets offer a promising alternative. They can broaden data accessibility, support reproducibility, and serve as a resource for algorithmic innovation, especially in rare and complex conditions, such as acute myeloid leukemia (AML) [4], [5]. Deep neural networks are known to be highly dependent on hyperparameter optimization (HPO) for achieving optimal performance across various tasks [6]. Recent research has started to extend this understanding to generative models for tabular data, showing that HPO can significantly impact the quality of generated synthetic datasets [7], [8]. However, investigations specifically focusing on HPO strategies for small and complex datasets, such as those encountered in clinical trials, remain scarce. Addressing this gap is essential for advancing synthetic data generation in clinical domains and other fields.

At the same time, the evaluation of synthetic data quality presents its own challenges, as no universally accepted methodology currently exists [9], [10], [11]. This lack of consensus leads to large variability in evaluation practices, as researchers often employ metrics tailored to their specific goals. However, this variability raises a critical question: if there is no standardized way to assess the quality of synthetic data, which metric should guide HPO? Furthermore, can a single metric adequately capture the diverse properties of synthetic datasets, or is a combination of metrics required? Existing studies highlight limitations in this regard. For example, Kotelnikov et al. [12] used a single machine learning prediction metric but did not measure improvements over default hyperparameters, whereas Kindji et al. [8] employed an XGBoost-based score, distinguishing real from synthetic data, for guiding HPO. Du and Li [7] combined one fidelity, one utility, and one privacy metric into a compound objective, however, none of these works compared different metrics within the same optimization framework. Consequently, the question of how to best guide HPO remains unresolved. Stoian et al. [13] underscored an additional challenge: generative models often violate domain-specific constraints, with some exceeding 95% non-compliance rates. These findings emphasize both the importance of optimizing hyperparameters and ensuring that models adhere to relevant domain constraints, a challenge particularly relevant to medical data.

We address these challenges by conducting a systematic comparison of HPO objectives for generative models tailored to synthetic clinical trial data. Building on our previous work [5], in which we introduced synthetic AML datasets, this study extends the scope by investigating an additional dataset, more generative models, incorporating a broader set of evaluation metrics, and exploring the impact of different optimization targets. We critically evaluate the absence of robust preprocessing and postprocessing steps, examine the ability of models to learn domain-specific constraints independently, and analyze the variability and interrelationships of evaluation metrics. Together, these analyses establish the foundation for identifying not only performance differences but also practical guidelines for applying generative models in clinical research.

In this study, we demonstrate that HPO with compound metric objectives, combined with robust pre- and post-processing, markedly improves the quality of synthetic clinical trial data. Our results highlight that systematic compound-metric HPO, explicit domain validation, careful metric selection, and dedicated privacy audits are all essential components for generating reliable and trustworthy synthetic datasets. These findings lead to actionable recommendations: prioritize multi-metric HPO strategies, integrate pre- and postprocessing steps to enforce domain-specific constraints, and include privacy auditing as a standard part of evaluation workflows.

## 2. Methods

### 2.1 Datasets

We used two clinical trial datasets:

1) **Acute Myeloid Leukemia (AML) Dataset**: This AML dataset consists of data collected from 1590 patients across four multicenter clinical trials [14], [15], [16], [17] and was already part of our previously published synthetic data generation pipeline [5], [18]. It includes 92 variables per patient, covering demographic, laboratory, molecular, and cytogenetic information, along with patient outcomes.

2) **ACTG320 Dataset (AIDS Clinical Trial)**: This publicly available dataset includes data from 1151 patients from an AIDS clinical trial [19] with 15 variables, including time-to-event data, treatment groups, and various patient characteristics such as age, sex, CD4 count, and prior medication use.

The dataset characteristics are depicted in Table 1. We limited our selection to two datasets to ensure a detailed and systematic evaluation while keeping computational demands manageable. While additional datasets were considered, a broader evaluation would have reduced the depth of analysis and increased complexity, limiting the clarity of our findings.

Both the AML and ACTG datasets were split into 80% training and 20% test sets, stratified by the combination of all binary outcome variables in each dataset. We retained missing values in the AML dataset rather than imputing them to better reflect realistic clinical conditions. In contrast, the ACTG dataset did not contain any missing values. In the AML dataset, all binary variables representing mutation status were transformed to the following format: 1 indicated a mutation, 0 indicated no mutation, and -1 indicated missing or unknown values (applicable to 13 mutations).

Based on initial experimentations with synthesizing both datasets, and in line with our previous work [5], we synthesized the difference between Event-free Survival Time (EFSTM) and Overall Survival Time (OSTM) ($EFSTM_{dif} = OSTM - EFSTM$) instead of synthesizing EFSTM directly. This approach better preserved the logical dependency between EFSTM and OSTM and reduced implausible cases where EFSTM would exceed OSTM, thereby producing more realistic survival data. After generating the synthetic dataset, we reconstructed EFSTM by applying $EFSTM = OSTM - EFSTM_{dif}$, restoring its original prior to computing any metric.

### 2.2 Generative Models

In this study, we evaluated nine generative models for synthesizing clinical trial datasets, using four different architectures: Generative Adversarial Networks (GANs), Variational Autoencoders (VAEs), Normalizing Flows (NFlows), and Denoising Diffusion Probabilistic Models (DDPMs). GAN-based models generate synthetic data by training a generator and discriminator in an adversarial setting to capture complex data distributions [20]. VAE-based models rely on an encoder-decoder architecture that learns a latent space representation of data and reconstructs samples from it [21]. Normalizing flows model probability distributions using a series of invertible transformations, allowing for flexible density estimation [22]. DDPM-based models iteratively learn to reverse a diffusion process that progressively adds noise to the data, enabling them to capture complex distributions of heterogeneous tabular features [12].

The selected models fall into two categories: general-purpose and survival-optimized models. A summary of the models evaluated in this study is provided in Table 2. To evaluate general-purpose models, we included CTGAN and TVAE, both widely used for tabular data synthesis but lacking explicit adaptations for survival data [23]. These models were used with their original implementations, which lack the automated pre- and postprocessing steps present in other models, such as those within the Synthcity framework [24]. Notably, these preprocessing steps include ensuring synthetic feature values remain within observed real data ranges. By using the original implementations, we aimed to assess the impact of missing preprocessing on the quality of synthetic data. Additionally, we evaluated CTAB-GAN+, a GAN-based model that incorporates robust pre- and postprocessing, improves the handling of rare categories and complex feature dependencies, and supports mixed-type variables [25]. We also included Tab DDPM, a diffusion-based model that leverages DDPM to handle both numerical and categorical features via Gaussian and multinomial diffusion processes [12]. Finally, we included RTVAE, a VAE-based model with enhanced robustness, to compare its performance in tabular data generation [26].

Beyond general-purpose models, we examined three survival-optimized approaches implemented within the Synthcity framework: Survival CTGAN, SURVAE, and Survival NFlow [24]. These models are not new architectures but rather adaptations of their respective base models, CTGAN, TVAE, and Normalizing Flows, integrated into Synthcity's Survival Pipeline. The pipeline modifies these models to handle time-to-event data but does not change their underlying generative structure. To ensure comparability with general-purpose models, we did not apply any additional censoring strategies beyond those present in the original data. While the three aforementioned models leverage existing architectures, Survival GAN represents a novel survival-specific GAN-based architecture that introduces novel loss functions and training mechanisms explicitly designed to model censored time-to-event distributions [27].

By evaluating both general-purpose and survival-optimized models, this study examined how generative architectures impact the quality of synthetic clinical trial data, guiding the selection of suitable data generation methods for clinical applications.

## 2.3 Data Quality Metrics

Evaluating synthetic tabular data remains challenging due to the absence of a universally accepted methodology [9], [10], [11]. Therefore, we selected multiple metrics for our evaluation, taking inspiration from the TabSynDex score [28]. We used four unmodified TabSynDex metrics, modified the fifth (Machine Learning Efficiency) for better alignment with real-world applications, and added two metrics to capture overall dataset similarity and survival data analysis. All metrics are summarized in Table 3.

We used the following metrics unmodified from the TabSynDex score:

**Basic Statistical Measure** compares means, medians, and standard deviations of numerical variables, computing and averaging relative errors.

**Regularized Support Coverage** measures how well the synthetic data reproduces the variable-level coverage of the real data, with particular emphasis on rare categories. We also include numerical variables in this comparison by converting them into ten bins.

**Log-transformed Correlation Score** evaluates the preservation of pairwise correlations using Pearson's correlation (continuous pairs), the correlation ratio (continuous–categorical), and Theil's U (categorical pairs), with a log transformation to lessen minor differences.

**$S_{pMSE}$ Index** evaluates how effectively a logistic regression model distinguishes between real and synthetic data. It refines the Propensity Mean Squared Error (pMSE) by comparing the observed pMSE to the expected pMSE ($pMSE_0$), where $pMSE_0$ represents the scenario in which synthetic data is completely indistinguishable from real data. The resulting ratio is then normalized using a factor alpha, ensuring the score lies within [0, 1]. Following the authors' recommendation, we used an alpha value of 1.2 [28].

Chundawat et al. used **Machine Learning (ML) Efficiency** as their fifth metric, which they defined as one minus the average relative error in predictive performance between models trained on synthetic data and those trained on real data [28]. Specifically, they used four models, logistic regression, random forest, decision tree, and a multi-layer perceptron. However, as argued by Kotelnikov et al. [12], comparing several suboptimal models is less meaningful than assessing the best model's performance. For tabular data, Gradient Boosting Tree methods typically yield superior results [29]. We decided to use CatBoost, which is superior to other Gradient Boosting Tree approaches when handling categorical data [30]. Similarly to Kotelnikov et al. [12], we conducted HPO for the CatBoost model on real data to simulate real-world usage. Since we predict only binary outcome variables, we use the Matthews correlation coefficient (MCC), which provides a balanced measure of predictive quality [31]. MCC is bound between -1 and 1. We used ML Efficiency in absolute terms so that its value reflects the synthetic data's performance independently of the baseline. This approach avoids instability when baseline performance is low and even allows for scores higher than those achieved on real data.

We noticed that the TabSynDex score lacks a specific metric to measure the overall similarity of the distribution of data points. To address this, we introduced the **K-Means Score**, inspired by Goncalves et al. [32], who used a similar approach to assess synthetic patient data quality. Unlike Goncalves et al. [32], who applied k-means clustering to the combined real and synthetic datasets, we first ran k-means (with k=10) solely on the real data to establish fixed centroid positions. This ensures that the evaluation is anchored to the true distribution of the real data, avoiding potential biases introduced by the synthetic data. We then used these centroids to cluster the synthetic data. For each cluster, we compared the proportion of synthetic samples to the corresponding proportion in the real data. To avoid situations where synthetic data heavily over-represents a single cluster and artificially inflates the overall score, we capped the maximum score for each cluster at 1. However, over-representation in one cluster naturally leads to under-representation in others, thereby lowering their individual scores. The K-Means Score is the average of all cluster-level scores. A perfect score of 1 indicates that every cluster contains the same proportion of real and synthetic data points.

Survival analysis is an essential part of the utility of clinical trial datasets. To evaluate how closely the synthetic data reflects real-world survival outcomes, we employ three metrics derived from Kaplan-Meier curve comparisons, as introduced by Norcliffe et al. [27]:

1. **Optimism** measures the discrepancy between the expected lifetimes in the synthetic and real data, quantifying a model's over-optimism or over-pessimism.
2. **Short-sightedness** quantifies the extent to which models trained on synthetic data fail to predict past a certain time horizon, hence capturing the temporal limitations in the synthetic data.
3. **Kaplan-Meier Divergence** represents the mean absolute difference between the synthetic and real Kaplan-Meier survival curves, measuring the overall match between the survival probabilities.

We rescaled these metrics so that 1 represents the best and 0 represents the worst possible value. Since all three metrics judge how close two Kaplan-Meier plots match, we use an average of these three metrics and call it the "**Survival Metric**".

All metrics, except ML Efficiency, which ranges from –1 to +1, were scaled from 0 (worst) to 1 (best). The evaluation of ML Efficiency depends on the specific prediction target used, meaning that different targets can lead to different overall utility assessments of the synthetic dataset. For the ACTG dataset, we experimented with using the Overall Survival (OS) Status and the Event Free Survival (EFS) Status as binary prediction targets (the underlying time-to-event variables, OSTM and EFSTM, are defined in Section 2.4). Since both performed rather weakly, we decided to use only the better-performing one for ML Efficiency, which was EFS. For the AML dataset, we decided to use OS and the first Complete Remission (CR1) as prediction targets. Consequently, we use seven metrics to evaluate the quality of synthetic ACTG datasets and eight metrics for synthetic AML datasets due to the inclusion of two prediction targets for ML Efficiency.

## 2.4 Domain Specific Validation

Ensuring that synthetic medical data adheres to domain-specific constraints is crucial for its validity and applicability in clinical research. Explicit clinical plausibility checks are necessary, as statistical similarity alone is insufficient to ensure that synthetic data aligns with real-world medical constraints and remains clinically meaningful [9], [33], [34].

To ensure that synthetic survival data preserves key clinical relationships, we defined a set of logical constraints that must hold in real-world survival data.

In survival analysis, two primary time-to-event variables must be considered:

- **Overall Survival Time (OSTM)** represents the total duration of survival from the start of the study until the last follow-up (either at the end of the study or when the patient leaves the study or is lost to follow-up [censored]) or the patient's death. OSTM must always be positive (OSTM > 0), as it measures the time until a definitive endpoint.

- **Event-Free Survival Time (EFSTM)** denotes the time from the start of the study until a specific event, such as relapse, progression, or death, occurs, or until the end of the study if no other event occurs first or until the patient is lost to follow-up (censored). Like OSTM, EFSTM must also be positive (EFSTM > 0) and cannot exceed OSTM (EFSTM ≤ OSTM).

Since EFSTM represents the first event occurring before or at OSTM, the proportion of cases where EFSTM = OSTM serves as an important validation measure. These instances correspond to cases where the first recorded event is either death or censoring at the same time point, reflecting the event structure in real-world survival data. Additionally, when EFSTM = OSTM, the overall survival status (OSSTAT) must match the event-free survival status (EFSSTAT) to ensure logical consistency. Beyond survival times, we also verified that numerical variables, such as age and clinical measurements (e.g., CD34, WBC), do not take unrealistic negative values in the synthetic data, mirroring the constraints of real-world datasets.

To assess whether synthetic survival data maintains domain-specific consistency, we tested the synthetic data for violation of the following logical constraints:

1. **OSTM > 0**
2. **EFSTM > 0**
3. **OSTM ≥ EFSTM**
4. **OSTM and EFSTM valid** (combination of 1, 2, and 3)
5. **If OSTM = EFSTM, then OSSTAT = EFSTAT**
6. **No Negative Values in Logical Variables** (Ensuring non-negative values for features like age, CD34, and WBC, except EFSTM and OSTM, which have separate constraints)
7. **Valid Patient Data** (combination of 4, 5, and 6)

To further evaluate the realism of synthetic survival data, we analyzed the proportion of cases where EFSTM = OSTM as a soft validation measure. This proportion serves as an indicator of how well the synthetic data preserves the event structure of real-world survival data.

We evaluated this proportion at two levels:

- **Exact match**: Cases where EFSTM and OSTM are identical.
- **Relaxed match**: Cases where EFSTM is within 95% of OSTM, allowing for minor discrepancies.

This relaxed comparison accounted for cases where EFSTM and OSTM were very close but not identical, reflecting small variations introduced by generative models. In postprocessing, it is possible to adjust these cases so that EFSTM = OSTM, aligning the synthetic data more closely with clinical expectations. However, in datasets such as AML, where 3% of real patients have EFSTM slightly lower than OSTM, adjusting the EFSTM value would remove a real patient subgroup when applied to synthetic patients. Therefore, evaluating both exact and relaxed matches provided insight into the models' ability to reproduce the real distribution, rather than enforcing an artificial correction. Ideally, the exact match proportion in the synthetic dataset should closely reflect that of the real data.

These validation checks are not used as optimization metrics but instead to assess the logical consistency and realism of the synthetic datasets. This helps us understand if the generative models can inherently learn these domain-specific constraints without explicit guidance.

## 2.5 Privacy Metrics

Synthetic data must be useful and non-disclosive. A trivial way to achieve high downstream utility would be to copy or near-copy training records. Conversely, generating samples far from the real data manifold would increase privacy but compromise utility. To verify that our generators neither memorize patients nor are systematically too close to the training set, we evaluate privacy using two distance-based metrics: Authenticity [35] and Adversarial Accuracy (AA)[36].

To reflect the predominance of binary/categorical variables in both datasets, all nearest-neighbor distances for both metrics are computed with Gower distance [37], rather than the Euclidean metric used in the original publications [35], [36]. Numerical features are min-max scaled using the training set only and categorical features (including binary) are treated as match/mismatch. Pairwise distances are then defined as the mean dissimilarity across observed features.

**Authenticity** is the fraction of synthetic records that are not unusually close to their nearest training neighbor, relative to how close real patients are to one another in the same neighborhood. For each synthetically generated data point $x_s$ we find its nearest real training record $x_r$ and compute the synthetic-to-real distance $d_{sr}$. For the same real data point $x_r$ we compute the leave-one-out nearest real-to-real distance $d_{rr}$. The data point $x_s$ is called authentic if $d_{sr} > d_{rr}$. Authenticity represents the fraction of authentic records in the synthetic dataset.

We investigate two additional summaries:

- **Distance ratio median (r median)**: for each synthetic data point $x_s$, we calculate $r = d_{sr} / d_{rr}$ and aggregate by median to summarize the typical local margin. Values >1 indicate a comfortable local margin from the matched training record (higher is better).

- **Too-Close 5% rate**: we compute the 5$^{th}$ percentile $\tau_5$ of the training leave-one-out real-to-real distances and report the fraction of synthetics with $d_{sr} < \tau_5$ (lower is better). This flags extreme proximity in the lower tail.

To calibrate expectations for non-training records, we apply the same procedure by treating the test set as if it were synthetic (test to train). This yields a hold-out baseline for Authenticity, r median, and Too-Close 5% rate that represents the proximity pattern of truly unseen real patients. In our experiments, we consider it a privacy concern when a synthetic generator produces lower Authenticity, smaller r median, or higher Too-Close rates than this hold-out baseline.

While Authenticity computes privacy at the record level, **AA** evaluates privacy at the set level, comparing cross-set versus within-set nearest-neighbor distances to see whether the synthetic distribution aligns more with the training set or with the held-out test set. AA is the average of two symmetric nearest-neighbor comparisons: for real points, the probability that their nearest synthetic neighbor is farther away than their nearest other real point (leave-one-out), and for synthetic points, the probability that their nearest real neighbor is farther away than their nearest other synthetic point. The value ranges from 0 to 1 and is approximately 0.5 when cross-set and within-set neighborhoods have comparable scale.

**Train AA** evaluates this between the real training data and the synthetic data. Values near 0.5 indicate that synthetic data points are neither unusually close to the training set nor systematically farther than training data points are to one another. Noticeably lower values point to potential overfitting, while higher values suggest separation from the training set and point to either underfitting or a distributional shift of synthetic data compared to the training data. **Test AA** evaluates the same calculation with the held-out testing set in place of train set. **Privacy loss** is defined as Test AA – Train AA. Positive values indicate that the synthetic dataset aligns more with the train than with the test set, which suggests overfitting and, thus, a privacy concern. Values near zero indicate no preferential alignment, while negative values typically reflect distributional shift away from the train set. This difference should be interpreted alongside the absolute AA levels: a small positive value when both AAs are well above 0.5 (e.g., ≈ 0.60) is a weaker signal of overfitting than the same difference when test AA ≈ 0.5 and train AA is clearly below 0.5. As a calibration, AA computed between the real training and test set is expected to be around 0.5. Deviations from this baseline reflect dataset-specific shift and help interpret absolute magnitudes.

We exclude Authenticity and AA from HPO because the privacy-utility trade-off is application-specific and cannot be sensibly pre-weighted. Adding privacy metrics to the objective risks sacrificing fidelity and downstream utility. Instead, we run synthetic data quality-first HPO and audit privacy post hoc against a hold-out baseline, identifying model-objective pairs that already meet privacy expectations and those that would warrant including explicit privacy metrics in the optimization objective or stronger regularization in future optimization work.

## 2.6 Hyperparameter Optimization

Recent studies show that HPO improves the performance of generative models [7], [8]. However, it remains unclear which optimization objective is most effective for synthetic data generation, particularly for clinical trial datasets. We define an HPO objective as the choice of evaluation metric or combination of metrics used as the objective function during optimization. Our goal is to compare different optimization objectives and quantify the improvement over default model configurations.

We evaluated two types of HPO approaches: single-metric optimization, where a single evaluation metric serves as the optimization target, and compound metric optimization, where multiple metrics are combined with equal weights into a single objective function. While an alternative approach was multi-objective optimization (MOO), which optimizes multiple objectives simultaneously without explicit weighting, we chose compound metric optimization due to the substantial computational overhead of MOO [38].

To assess the impact of different objective functions, we tested four optimization objectives:

- **ML Objective** (single-metric optimization): Used only the ML Efficiency metric as the optimization target. For AML, we optimized for OS due to its greater clinical importance.
- **Survival Objective** (single-metric optimization): Used only the Survival Metric as the optimization target.
- **Four Metrics Objective** (compound metric optimization): Combined ML Efficiency, Survival Metric, $S_{pMSE}$ Index, and Log-Transformed Correlation Score with equal weights. For AML, ML Efficiency was based on OS.
- **Full Objective** (compound metric optimization): Combined all evaluation metrics (seven for ACTG, eight for AML) with equal weights.

All metrics, except ML Efficiency (which ranges from –1 to +1), were scaled between 0 (worst) and 1 (best) to ensure comparability across the compound metric optimization objectives.

For HPO, we used Optuna [39] with the Tree-structured Parzen Estimator (TPE) Sampler [40], conducting 30 optimization trials per objective for each generative model. Given the small dataset sizes, we used five-fold cross-validation instead of a separate validation set. Each trial consisted of five rounds, where the generative model was trained using data from four of the five cross-validation folds, and synthetic data was sampled to match the training data size. Metrics were computed according to the respective optimization objective, and this process was repeated across all cross-validation sets. The final trial score is the average across these five runs. Figure 1 provides an overview of the entire HPO procedure.

To improve efficiency, we applied early stopping to discard less promising hyperparameter configurations. After each round, the current average score was compared to the best score observed so far. If a trial's score was at least 10% lower than the best-completed trial, further rounds were not conducted, and the current score was returned.

Additionally, if a generative model produced invalid outputs (e.g., null values) for a given hyperparameter configuration, the trial was immediately discarded and assigned a score of zero.

To ensure comparability across optimization objectives, we fixed the training and sampling seeds for all generative models, and a consistent random seed was used throughout the optimization process. No parallelization was applied to maintain identical starting conditions across different optimization targets. After 30 trials per generative model, the best-performing hyperparameter configuration for each objective was saved.

For the five generative models implemented within the Synthcity framework, we used the pre-defined hyperparameter spaces provided by the framework. For the remaining four models, the hyperparameter search spaces are displayed in Table A1.

All experiments were conducted on a workstation equipped with an Intel Core i9-13900K CPU, 64GB RAM, and an NVIDIA RTX 4090 GPU.

## 2.7 Experimental Design

We trained the nine generative models with five different sets of hyperparameters: the default set and four derived from the HPO objectives*: Survival Objective*, *ML Objective*, *Four Metrics Objective*, and *Full Objective*, using the original 80:20 training-test split. To mitigate the impact of randomness and analyze the stability of the models, each model was trained five times for each hyperparameter configuration using five different random seeds. Additionally, we used five distinct random seeds for sampling the synthetic data. This process yielded 25 synthetic datasets per hyperparameter set, resulting in a total of 225 generative models and 1125 synthetic datasets for evaluation. Each synthetic dataset had the same size as the training set.

The evaluation process, as illustrated in Figure 2, was organized into several key steps. In Step 1, we compared the performance of the four HPO objectives based on the average of all data quality evaluation metrics described in Section 2.3. This comparison quantified the improvement over default hyperparameters and identified the best-performing objective for each generative model. Additionally, we evaluated whether specific objectives outperformed others relative to their respective optimization goals. To complement these analyses, we also performed an initial privacy assessment to explore how different optimization objectives affected privacy relative to synthetic data quality. To assess the efficiency of the four HPO objectives, we monitored the optimization duration for each model.

Once the best-performing hyperparameter objective was identified, we conducted in Step 2 a detailed model evaluation to compare the individual metrics for all generative models. In this step, we performed a more detailed privacy analysis using all metrics described in Section 2.5. This allowed us to assess privacy risks at the model level in greater depth and relate them directly to synthetic data quality. We also analyzed the performance of general-purpose models versus survival-specific models. Additionally, we directly compared CTGAN with its survival-specific variant (Survival CTGAN) to assess the effectiveness of survival-oriented optimization. To provide a reference for comparison, we applied the same data quality evaluation metrics to real data, treating the training data as if it were synthetic and using the test data as the ground truth. Note that this comparison was not entirely fair, as the training and test data were not identical in size: the training data consisted of 80% of the total data, while the test data comprised just 20%. Therefore, it was not expected that these values represent the strict upper bound for all the metrics. However, they served as a reference point for what might be expected from high-quality synthetic datasets.

To further validate the models, we examined in Step 3 how well the domain-specific constraints described in Section 2.4 were preserved in the synthetic datasets, aiming to determine if the models could learn these constraints independently and whether survival-specific models violated fewer constraints. To assess the impact of preprocessing, we compared generative models with robust pre- and postprocessing to models without it. Then, we reverted the changes described in Section 2.1 and synthesized EFSTM directly instead of using the difference between OSTM and EFSTM to explore the impact of preprocessing on the dataset level. We therefore trained another 225 generative models using the same hyperparameters and training seeds and generated another 1125 synthetic datasets using the same sampling seeds. Following the domain validation, we performed a reevaluation of the models (Step 4) after removing invalid data points that violated our defined constraints. We compared the individual metrics for each model with the achieved results before the removal to investigate which of the metrics benefitted from the removal and which did not.

Finally, in Step 5, we evaluated the variability of individual data quality metrics across the 1125 synthetic datasets (originated from Step 1) by analyzing their ranges and standard deviations. This analysis aimed to identify patterns in metric behavior across both datasets, providing insights into their responsiveness to changes in synthetic data quality. To complement this, we analyzed inter-metric correlations to detect potential redundancies, ensuring that metrics used in optimization objectives capture diverse aspects of data quality. These evaluations were conducted to better understand the relative stability, sensitivity, and independence of individual metrics, guiding their use in optimization and evaluation frameworks.

In conclusion, the insights gained from these experiments allowed us to derive actionable recommendations for optimizing hyperparameters of generative models in order to generate high-quality synthetic datasets.

# 3. Results & Discussion

## 3.1. Hyperparameter Optimization

We evaluated four optimization objectives: *ML Objective*, *Survival Objective*, *Four Metrics Objective*, and *Full Objective*, against the default hyperparameters for the ACTG and AML datasets. The optimization runs for all nine models combined required between almost 19 hours (*ML Objective*) and 60 hours (*Four Metrics Objective*) on the ACTG dataset and between 127 hours (*Survival Objective*) and 232 hours (*Full Objective*) on the AML dataset (Table A2). Notably, we did not set any time limitations for the trials or use parallel trials, to ensure comparability across optimization objectives. More detailed information on optimization times, including variations across models, can be found in Table A3.

Given the substantial time requirements, focusing on a single optimization objective is more practical in real-world scenarios, making it essential to identify the most effective approach. To assess the effectiveness of these objectives, we computed the average of all chosen data quality evaluation metrics for the 25 synthetic datasets generated for each model and hyperparameter set. Additionally, we ranked the objectives according to their average performance for each model and calculated the average rank for each objective across all models.

Overall, the objectives *Four Metrics Objective* and *Full Objective* yielded the best results, with average ranks of 2.00 and 1.78 on the ACTG dataset and 1.78 and 1.89 on the AML dataset, respectively (Table 4). In contrast, the models with default hyperparameters performed the worst, with average ranks of 4.11 on the ACTG dataset and 4.89 on the AML dataset. The average improvement over models with default hyperparameters on the ACTG dataset ranged from 3% for the *ML Objective* to 19% for the *Full Objective*. On the AML dataset, the average improvement ranged from 11% for the *Survival Objective* to 46% for the *Full Objective*. However, these large relative gains on AML are mostly driven by Tab DDPM, which exhibited a particularly strong improvement due to weak baseline performance under default hyperparameters. When excluding Tab DDPM, the improvements on AML decrease notably, ranging from 11% for the *Survival Objective* to 23% for the *Four Metrics Objective*.

These findings show the general advantage of compound metric optimization objectives, which appear to be well-suited for producing synthetic datasets that balance multiple evaluation goals. While the percentage improvements achieved through HPO might seem modest, they could still be meaningful in practice. In scenarios where synthetic datasets are used for downstream analyses, even moderate gains in evaluation metrics may help the synthetic data cross critical thresholds of similarity to real data, potentially enabling their use in tasks such as comparative effectiveness analyses.

The observed performance improvements varied significantly between models, as shown in Table 4. Tab DDPM achieved by far the largest relative gains on the AML dataset, where its default hyperparameters resulted in a very low baseline performance of 0.23, compared to an overall average of 0.54 across models. Consequently, compound optimization objectives improved its performance by more than 300%. In contrast, improvements on the ACTG dataset were more modest, of up to 38%. TVAE achieved substantial enhancements on both datasets, improving by up to 53% on ACTG and 60% on AML, starting from relatively low default scores of 0.51 and 0.47, respectively. CTGAN and CTAB-GAN+ also benefited considerably, with improvements of around 20-25% on the ACTG datasets and of up to 38-39% on the AML dataset. Notably, these four models were the only models for which we used the original implementations and not the implementations provided by Synthcity. This highlights that their default hyperparameters are not well suited for small datasets, making HPO particularly beneficial.

Certain configurations led to worse performance for certain models. For example, hyperparameter optimization was not beneficial for Survival GAN on the ACTG dataset. Additionally, other models experienced decreased performance when optimized for a single metric on the ACTG dataset: Tab DDPM exhibited the largest drop with 36.9% under the *ML Objective*, followed by Survival CTGAN with an 11.9% decline under the ML Objective, while CTGAN (3.5%) and SURVAE (2.0%) showed smaller, non-significant decreases. On the AML dataset, only the SURVAE model showed a 6.8% decrease in performance under the *ML Objective*. These observations highlight that while rare, single-metric optimization objectives can sometimes lead to overfitting or performance imbalances, making compound metric optimization a more reliable choice for consistent improvements.

We investigated whether the optimization objectives, even if they did not achieve the best average metric results overall, might excel in the specific metric or combination of metrics they were optimized for. As shown in Table 5, *Four Metrics Objective* and *Full Objective* consistently performed best across all evaluation criteria on the ACTG dataset, with the exception of *Survival Objective* slightly outperforming *Full Objective* for its specific evaluation. On the AML dataset, single-metric optimization targets, *ML Ob*jective and *Survival Objective*, performed best for most models in their respective metric evaluation, suggesting some benefit in optimizing for a single metric. However, despite these localized improvements, their average scores across all models were not the highest. Additionally, this approach limits the dataset's broader usability, as their results in other metrics were lower than those achieved by compound metric optimization objectives. *Four Metrics Objective* and *Full Objective* performed similarly across the evaluation of metric assemblies, showing only small differences between them. For both datasets, the performance gap between these compound objectives and the single-metric approaches was substantial.

To complement these analyses, we evaluated the privacy of the generative models. Initial observations suggested that privacy outcomes depended more strongly on the model architecture than on the chosen optimization objective. To investigate this systematically, we quantified the variability in privacy metrics across models and objectives by comparing the median standard deviation across objectives per model with the median standard deviation across models per objective. On the ACTG dataset, variability across models was 1.62 times higher than across objectives for Authenticity and 2.63 times higher for Privacy Loss. On the AML dataset, the effect was even more pronounced for Authenticity, where variability across models was 3.10 times higher than across objectives, while Privacy Loss ratio was 1.98. These results indicate that, while optimization objectives influence performance, the inherent characteristics of the generative models have a stronger effect on privacy outcomes.

Motivated by this finding, we further investigated the interplay between privacy and synthetic data quality at the model level. Figures 3 and A1 show Adversarial Accuracy (AA) and Authenticity in relation to synthetic data quality, allowing us to contextualize privacy behavior in terms of the overall generative performance. We observed several notable patterns. First, changes in Authenticity were generally more pronounced than variations in AA, indicating that Authenticity is the more sensitive privacy measure in this context. Second, several models, such as RTVAE and Survival GAN consistently showed high AA across all configurations and both datasets, suggesting underfitting. Third, TVAE exhibited persistent privacy concerns: across all configurations and on both datasets, Authenticity dropped below the test set baseline, even under default hyperparameters where synthetic data quality was low. While Privacy Loss and Authenticity typically moved in the same direction, TVAE demonstrates that high Train and Test AA does not necessarily imply acceptable privacy, highlighting the need to consider both metrics jointly. Fourth, Privacy Loss was rarely positive overall, indicating that synthetic data were not systematically closer to the training set than to the test set. Finally, we observed that improvements in synthetic data quality did not universally come at the cost of privacy. For several models, such as SURVAE on ACTG and Survival CTGAN on AML, configurations achieving higher data quality exhibited not only comparable but even higher Authenticity values, suggesting lower memorization risk. Moreover, for some models, comparable data quality could be reached under different optimization objectives while achieving notably better privacy.

In summary, while single-metric objectives can provide targeted improvements, particularly on the AML dataset, their limited generalizability reduce their practical utility. By contrast, the compound objectives deliver stronger and more consistent results, underscoring their importance for achieving balanced performance, despite their higher computational costs. Privacy outcomes, however, were found to depend more on the model architecture than on the optimization objective. For some models, better-performing optimization objectives simultaneously achieved higher data quality and better privacy, indicating that better data quality does not always require sacrificing privacy.

## 3.2 Model Evaluation

To obtain a better understanding of the generative models and their capabilities, we compared them regarding our chosen metrics. For a fair comparison, we used only the best-performing hyperparameter configuration per model, selected based on the average of all chosen data quality metrics. We present the average and standard deviation of the individual metrics for the 25 synthetic datasets for each model in Figure 4. Additionally, we compared the results with the real data itself by treating the training data as if it were synthetic and using the test data as the ground truth for the calculation of metrics, providing a benchmark for expected performance from high-quality synthetic datasets.

On both datasets, the general-purpose models TVAE and Tab DDPM performed best across most metrics. While achieving similar average performance, their strengths differed across individual metrics. TVAE achieved particularly strong results on the Log-transformed Correlation Score and performed better on variable-level fidelity metrics: Basic Statistical Measure and Regularized Support Coverage. In contrast, Tab DDPM showed clear advantages in utility metrics, including ML Efficiency and the Survival Metric, and also achieved higher K-Means Scores. The next best model on both datasets was Survival CTGAN, however with quite a gap in the performance to both models, especially on the ACTG dataset. RTVAE performed poorly on both datasets. On the ACTG dataset, however, Survival GAN, which did not benefit from HPO, performed even worse. On the AML dataset, Survival GAN performed better, achieving average performance.

Looking at individual metrics, we find that using the real training-versus-test comparison as an approximate upper bound is informative but not universally reliable. For several metrics, including Basic Statistical Measure, ML Efficiency, Survival Metric, and the $S_{pMSE}$ Index, the real-data baseline provided a good reference point, with only a few models surpassing it. In contrast, the Log-transformed Correlation Score failed as a meaningful benchmark: six out of nine models exceeded the real-data score on ACTG, and all models exceeded it on AML. This likely results from distributional mismatches when comparing 20% of a small dataset to the full 80%, where most variable pairs have near-zero correlation. While the logarithmic transformation reduces this effect, it does not fully resolve it. In future work, introducing a minimum correlation threshold (e.g., skipping variable pairs with correlations < 0.1) could yield more interpretable reference values. Despite these limitations, such real-data baselines remain highly valuable, especially in practice, as they allow quick assessments of whether generated data reaches expected quality levels.

To better understand the trade-offs between utility and privacy, we further analyzed privacy outcomes across models (Figure 5). For context, we compared them to the hold-out baseline derived from test-to-train comparisons. This baseline indicates how close unseen real patients can naturally lie to the training set and thus provides a reference for assessing memorization risk.

On ACTG, the two best-performing models, TVAE (0.791) and Tab DDPM (0.790), showed substantial privacy concerns. Both exhibited a strong privacy loss (≈0.07-0.08) and substantially lower Authenticity scores with 0.324 and 0.351 compared to the 0.459 for the test-set baseline, indicating overfitting. While both models also produced more samples in the lower-distance tail, TVAE was particularly problematic: almost 18% of its synthetic records fell below the Too-Close 5% threshold, compared to 10% for Tab DDPM. In contrast, Survival CTGAN and CTAB-GAN+ achieved considerably better privacy results with only a moderate reduction in data quality (0.763, and 0.756, respectively).

On AML, TVAE showed even more problematic results across all privacy metrics: Authenticity dropped to 0.387 compared to the much higher test-set baseline of 0.723, the Too-Close 5% rate reached 21%, the median distance ratio was substantially lower at 0.853 compared to 1.356 for the baseline, and Privacy Loss was extremely elevated (0.169), indicating strong overfitting. Tab DDPM performed considerably better, achieving an Authenticity of 0.697 and an r median of 1.304, but still falling slightly below the expected privacy thresholds and exhibited a small positive Privacy Loss (0.026). However, its Train AA remained around 0.5, suggesting no strong overfitting, making it a more acceptable, but still somewhat riskier, option when prioritizing data quality. Survival CTGAN (0.751) and CTAB-GAN+ (0.748) represented again safer alternatives, offering better privacy while achieving slightly lower synthetic data quality compared to Tab DDPM (0.766) and TVAE (0.761).

Interestingly, poor generative performance did not necessarily imply safe privacy behavior. On AML, SURVAE, despite being among the lower-performing models, still showed notable privacy concern. Similarly, on ACTG, Survival GAN, while producing the lowest-quality synthetic data, exhibited elevated memorization risk.

Finally, the observed differences in expected Authenticity and r median between ACTG and AML primarily arise from dataset characteristics: AML contains substantially more variables, which naturally increases typical distances between samples, resulting in higher baseline values. Importantly, because both datasets are relatively small, not all synthetic records can lie comfortably outside the training distribution, even when privacy is well preserved. This highlights the value of reference baselines: while Alaa et al. [35] suggested discarding unauthentic samples post hoc to improve privacy and utility, our results indicate that, in our setting, such samples may still correspond to real unseen patients and should therefore not be removed.

Considering both data quality and privacy, Survival CTGAN emerged as one of the best-balanced models in our study. It also noticeably outperformed CTGAN on both datasets. However, since we used the original implementation of the CTGAN model, the implementation details differ. The Synthcity framework provides automatic pre- and postprocessing, and the hyperparameter options between the two implementations vary, making it impossible to exactly match the hyperparameters. Consequently, the optimizations of both models were independent of each other (different hyperparameter spaces), making a fair comparison challenging. To address this, we used the Synthcity implementation for both models in a controlled comparison. We used identical hyperparameters, training procedures, and sampling seeds so that the only difference between them was the way the training data was fed to the network. The results displayed in Figure A2 show that when matched with the same hyperparameters, CTGAN overall outperformed Survival CTGAN. This finding was surprising, given that Survival CTGAN already achieved strong results.

So far, our study found no consistent evidence that survival-optimized models are superior to general-purpose models. This suggests that when synthesizing clinical trials, it may be insufficient to rely solely on models optimized for specific tasks. Instead, comparisons should include general-purpose models, such as Tab DDPM and CTAB-GAN+, as they might outperform survival-optimized models.

## 3.3 Domain Specific Validation

Beyond overall data quality and privacy, synthetic datasets must also adhere to domain-specific clinical constraints to be useful in practice. To examine this aspect, we evaluated the validity of the generated synthetic datasets as outlined in Methods Section 2.4. Patients violating any of the defined logical constraints (*V*1-*V*7) were classified as faulty. Additionally, we compared the ratio of patients with matching OSTM and EFSTM times, examining exact matches and relaxed matches (within 5% tolerance).

The evaluation revealed significant differences in the proportion of faulty patients across models, as shown in Figure 6. TVAE and CTGAN, which lack robust pre- and postprocessing, exhibited the highest violation rates on the AML dataset. On the ACTG dataset, however, this was not the case for TVAE. We attribute this to the dataset's simpler structure, with fewer variables and key variables (e.g., OSTM and EFSTM) represented as integers rather than floats, making it easier for models to learn their distributions.

The lack of robust pre- and postprocessing in these two models led to the generation of negative values, which resulted in the exclusive violations *V1* (OSTM < 0), *V3* (OSTM < EFSTM), and *V6* (other negative values). Note, that EFSTM is not directly synthesized but instead derived from $EFSTM_{dif}$, subtracted from OSTM. A negative $EFSTM_{dif}$ value results in patients with EFSTM exceeding OSTM times, which is medically implausible. Violations of *V3* (OSTM <

EFSTM) and *V2* (EFSTM < 0) were particularly frequent in these two models. In contrast, models with robust pre- and postprocessing, such as SURVAE and Survival CTGAN, demonstrated stronger adherence to logical constraints. On the ACTG dataset, models with robust pre- and postprocessing had fault rates (*V7*) between 3% and 11%, compared to 44% for CTGAN. On the AML dataset, these models achieved fault rates below 30%, while TVAE and CTGAN exhibited higher rates of 41% and 61%, respectively. Survival-optimized models generated noticeably fewer faulty patients, particularly on the AML dataset, averaging under 10% faulty patients compared to the best general-purpose model, which still generated double the faulty patients. This finding suggests that although survival models may not excel in overall metrics, they produce synthetic data that aligns closer to clinical expectations.

Regarding matching OSTM and EFSTM times, most models closely replicated the original ratio of 93% on the ACTG dataset, with the notable exception of CTGAN, which generated only about one-third of the required ratio. In contrast, on the AML dataset, all models except for CTAB-GAN+ struggled to replicate the real ratio, particularly in exact matches. Interestingly, Tab DDPM was the second-best model at replicating this ratio and the only one to generate more patients with matching OSTM and EFSTM times (52.6%) than the real ratio of 44.5%. The two models without robust pre- and postprocessing, CTGAN and TVAE, performed worst, failing to generate a single patient with matching times under exact evaluation. Even under the relaxed evaluation (5% tolerance), their proportions improved only to 16% and 18%, respectively, which remain far behind all other models. CTAB-GAN+ stood out as the only model that achieved consistently good results in replicating matching ratios. Its success can be attributed to its ability to generate mixed variables. This feature allows it to treat $EFSTM_{dif}$ as a categorical variable (e.g., 0 for non-existent values) and, when applicable, generate numeric outputs for the remainder. Combined with our transformation of the original EFSTM variable, this capability enabled CTAB-GAN+ to achieve matching ratios close to the real data.

As observed, pre- and postprocessing of generative models significantly reduced violations in synthetic data. To further quantify this impact at the dataset level, we reversed the EFSTM transformation and instead synthesized EFSTM values in their original form (Figure 7). We used the same hyperparameters and seeds for the comparison. Removing the EFSTM transformation increased the proportion of faulty patients across both datasets. On the ACTG dataset, the average fault rate rose from 10% with the transformation to 48% without it, primarily due to EFSTM exceeding OSTM (*V3*). On the AML dataset the impact was considerably smaller, fault rates increased from 23.25% to 26.49%. These increases, nevertheless, highlight the critical role preprocessing plays in ensuring logical consistency.

The EFSTM transformation also substantially influenced the proportion of patients with matching OSTM and EFSTM times. On the ACTG dataset, relaxed match proportions dropped from 88% with the transformation to an average of 31% without it. Exact matches showed an even larger contrast: TVAE achieved only 3% exact matches without the transformation but improved to 93% when it was applied. On the AML dataset, relaxed match proportions decreased from 35% with the transformation to just 11% without it. Tab DDPM was an exception, achieving 42.6% matching patients under the relaxed condition and 15.9% for exact matches, performing better than all other models on AML.

Nevertheless, the results demonstrate overall the importance of the EFSTM transformation in supporting models to replicate real data distributions.

This analysis highlights the importance of robust pre- and postprocessing in generating logically consistent synthetic data. Models lacking these steps, such as TVAE and CTGAN, exhibited significantly higher violation rates and struggled to replicate real data distributions more. Additionally, survival-optimized models consistently generated fewer faulty patients, demonstrating their clinical relevance despite not always achieving the best performance on general metrics. The EFSTM transformation proved critical for mitigating logical inconsistencies, particularly for violations involving EFSTM exceeding OSTM. Removing the transformation led to substantial increases in fault rates and reduced the ability of models to replicate the correct distribution of EFSTM = OSTM times. To improve the quality of synthetic survival datasets, we recommend prioritizing domain-specific preprocessing strategies like the EFSTM transformation and ensuring that models are equipped with robust pre- and postprocessing mechanisms. These steps are essential for achieving logically consistent, high-quality synthetic datasets that align with real-world data.

## 3.4. Model Reevaluation

As the next step, we removed all non-valid patients ($V7$) from the 25 synthetic datasets generated by each model using their best HPO objective. This reevaluation assessed how removing non-valid patients influenced performance metrics. On the ACTG dataset, an average of 7% of patients were removed, ranging from 2% for TVAE to 26% for CTGAN. For the AML dataset, the proportion was significantly higher, with an average of 22% removed, ranging from 8% for SURVAE to 61% for CTGAN. While the removal of faulty patients generally led to decreases in average metrics, the impact was smaller than expected, with average reductions of 0.0052 on ACTG and 0.0090 on AML. Interestingly, CTGAN on ACTG and RTVAE on AML were exceptions, showing improvements after patient removal.

Table 6 summarizes the changes for each metric and dataset, showing the number of models that benefited from the removal and the average differences. While Basic Statistical Measure, Log-transformed Correlation Score, Regularized Support Coverage, K-Means Score, and ML Efficiency showed consistent declines after patient removal across most models, these reductions were generally not statistically significant. The only exceptions were Regularized Support Coverage on ACTG (-0.0176) and ML Efficiency (CR1) on AML (-0.0228), which showed significant decreases. $S_{pMSE}$ Index was the only metric that on average improved across both datasets, with gains of 0.0033 on ACTG and 0.0225 on AML, though these changes were not statistically significant.

Since general-purpose models exhibited more domain violations, patient removal had a greater impact on their metrics than on survival-optimized models, particularly on the AML dataset (Figure A3). The largest decreases in average metrics were observed for CTGAN and TVAE, with declines of 0.04 and 0.02, respectively. After removal, Survival CTGAN (0.7505) outperformed TVAE (0.7400) in the average of metrics, becoming the second-best performing model for AML after Tab DDPM (0.7544). Nevertheless, even after the removal of 40% patients, TVAE still ranked third, demonstrating that substantial patient removal does not make synthetic datasets unusable.

Overall, while the removal of non-valid patients led to slight declines in most metrics on average, the majority of these changes were not statistically significant. This suggests that generating more patients than necessary and subsequently applying postprocessing to remove faulty records can still be a viable strategy without substantially compromising overall data quality. However, robust pre- and postprocessing mechanisms remain essential for minimizing domain violations. Survival-optimized models, which generated fewer faulty patients, showed smaller metric variations after removal, underscoring their robustness. Nevertheless, survival-optimized models did not consistently outperform general-purpose models in overall metrics, even after patient removal.

Future efforts to generate high-quality synthetic clinical trials should include a comparison of both model types, but this evaluation must occur after removing patients that violate critical constraints. For general-purpose models, which tend to violate more constraints, it is particularly important to consider the implications of patient removal. The more patients that are removed during postprocessing, the more synthetic data must be generated initially to compensate, which increases the risk of distributional shifts in the synthetic data. These shifts can reduce the alignment with the real data and compromise the overall stability of the generation process. To mitigate these issues, prioritizing robust preprocessing strategies at both the model and dataset levels is essential for generating logically consistent and high-quality synthetic datasets.

## 3.5 Metric Evaluation

To better understand the metrics used for evaluating synthetic data quality, we analyzed their behavior across all 1125 synthetic datasets. This analysis aimed to assess the suitability of individual metrics for optimization.

Figure A4 illustrates the distributions of metrics for default and optimized configurations across synthetic datasets for ACTG and AML. Metrics generally exhibited broader distributions in the AML dataset, reflecting its higher complexity and dimensionality. Note that several Tab DDPM configurations, specifically the *ML Objective* on ACTG and the default and *Survival Objective* on AML, performed substantially worse than all other models. These outlier configurations, visible as long tails in Figure A4, inflated the reported standard deviations for most metrics but are not representative of the general behavior across models.

Survival Metric showed narrower ranges than other metrics, suggesting stability, but limited responsiveness to changes in synthetic data quality. Conversely, metrics like the $S_{pMSE}$ Index and Log-transformed Correlation Score demonstrated wider ranges, indicating higher sensitivity to variations in synthetic data quality and greater potential for optimization. Optimization reduced metric variability in most cases, particularly for the AML dataset. However, some metrics, such as the Survival Metric, displayed minimal differences between default and optimized configurations on ACTG, suggesting limited responsiveness to optimization. This highlights that metrics with wider distributions and higher variability, such as the $S_{pMSE}$ Index, may offer greater utility in guiding optimization processes. Interestingly, the broader ranges and variability of default configurations could serve as proxies for identifying metrics that are more sensitive to data quality changes. Future studies should explore the effectiveness of default variability and spread as a predictor of optimization outcomes.

Figure A5 presents the correlation matrices for metrics across both datasets. Certain metrics, including the Basic Statistical Measure, K-Means Score, Log-transformed Correlation Score, and $S_{pMSE}$ Index, exhibited consistently strong correlations, suggesting potential redundancy in evaluation. These metrics appear to capture overlapping aspects of synthetic data quality and may require careful weighting in compound optimization objectives in the future to avoid overemphasizing related features.

In contrast, metrics like the Survival Metric and ML Efficiency scores showed weaker correlations with others, indicating that they capture more independent characteristics. However, in the case of the ACTG dataset, the ML Efficiency showed at most very weak correlations with all metrics, which, when combined with the low MCC score of 0.1221 on the original dataset, suggests that this metric is unstable in this dataset. This is not the case on the AML dataset, where the MCC values on the original data showed moderate (OS) and strong (CR1) performance. These findings underscore that metric selection should be dataset-dependent, balancing stability and responsiveness while ensuring that redundant metrics do not dominate compound optimization objectives. Additionally, the weak correlation between standard metrics and domain-specific validity reinforces the need for explicit clinical plausibility checks rather than relying solely on statistical similarity and utility.

In summary, metrics such as the $S_{pMSE}$ Index and Log-transformed Correlation Score are particularly useful for optimization due to their high variability, whereas more stable metrics, such as the Survival Metric, provide robustness but may be less informative for guiding optimization. Careful weighting is necessary to balance redundancy in compound metrics. Additionally, weak correlations between standard metrics and domain-specific validity highlight the importance of explicit validation steps to ensure clinical relevance. Future work should explore how the distribution of metric values in default configurations can be leveraged to refine optimization targets.

## 3.6 Overall Discussion

To the best of our knowledge, this is the first study to systematically compare multiple HPO objectives for synthetic tabular data generation. A key consideration in our selection of HPO objectives was the computational cost, limiting our ability to explore an even wider range of methods. The single-metric optimization objectives we employed were utility-driven: *ML Objective* followed Kotelnikov et al.'s [12] approach, using ML Efficiency, while *Survival Objective* focused on survival analysis metrics, which were specifically designed for clinical trial datasets [27]. In contrast, the compound metric optimization objectives, *Four Metrics Objective* and *Full Objective*, integrated these metrics into broader evaluation criteria, resulting in a more balanced performance. Although compound metric objectives outperformed single-metric approaches in our experiments, the K-Means Score showed the highest correlations with all other metrics. Exploring it as a single-metric optimization target could therefore bridge the gap, offering a more holistic objective for synthetic data generation.

Our results demonstrated that while all evaluated generative models benefited from HPO, improvements varied across models. Tab DDPM improved by far the most, due to poor performance with default hyperparameters. CTGAN and TVAE showed the second-largest enhancements, aligning with findings by Kindji et al. [8], who similarly observed substantial improvements for these two models, suggesting their default hyperparameters are particularly suboptimal. In contrast, Du and Li [7] reported smaller improvements from HPO, likely due to the inclusion of privacy metrics in their optimization process. This aligns with the known trade-off between privacy and utility [41], [42]. Nevertheless, our findings indicate that better privacy outcomes for the same generative model do not necessarily imply worse fidelity or utility. However, the question of how to best integrate privacy metrics and balance them against other objectives without compromising synthetic data quality remains open. In addition, synthetic datasets generated using default hyperparameters exhibited similar metric ranges and variances comparable to optimized datasets. Future research should explore how leveraging metrics from default models, in combination with metrics computed on real data, can guide the selection of evaluation metrics a priori and inform more effective optimization objectives.

While HPO improves synthetic data quality, ensuring domain-specific consistency remains essential for generating clinically valid datasets [9], [33], [34]. Notably, none of the evaluated generative models inherently learned to adhere to domain-specific clinical constraints, reinforcing the need for explicit validation steps. This is consistent with the observations of Stoian et al., who reported non-compliance rates exceeding 95% for some models [13]. Importantly, there was no clear correlation between evaluation metrics and the proportion of invalid patients. This finding highlights the limitations of conventional evaluation metrics, which fail to account for fundamental domain constraints. Therefore, future evaluations should explicitly account for domain violations by analyzing model performance post removal of invalid synthetic records, providing more accurate and clinically relevant assessments. These findings strongly suggest that ensuring domain consistency requires explicit integration of domain knowledge into the synthetic data generation pipeline, rather than relying solely on HPO or standard evaluation metrics.

To enforce adherence to constraints, some frameworks provide a way to define explicit rules during data generation [13], [24], [43]. However, these frameworks have limitations, such as supporting only specific constraint types or being restricted to compatible generative models. Notably, Synthcity [24] stands out due to their comprehensive pre- and postprocessing capabilities and broad model compatibility. As an alternative or complementary approach, post-hoc removal of invalid synthetic data could be viable. Our results show that removing up to 60% of faulty patients caused only moderate metric declines, suggesting this strategy's feasibility. However, since our evaluation focused on relatively simple constraints, future research should reassess this approach with more complex constraints. Additionally, post-hoc removal requires generating more synthetic data than needed initially, and the removal of non-random faulty data risks introducing distribution shifts, potentially destabilizing the generation process. Therefore, while useful, this approach should be applied cautiously, complementing rather than replacing robust preprocessing and postprocessing strategies.

Our privacy evaluation showed that the choice of a specific model architecture had a bigger impact on privacy outcomes than a specific HPO objective. Across both datasets, TVAE demonstrated the most concerning privacy behavior, consistently showing low Authenticity and high rates of synthetic samples falling within the "too-close" region, even when using default hyperparameters where the overall synthetic data quality was low. Importantly, Authenticity scores were not particularly high even for the real-data baselines obtained from test-to-train comparisons. This observation is somewhat expected given the relatively small sample sizes of our datasets, where truly unseen patients naturally lie closer to the training set than would be typical in larger datasets. While this baseline provides a useful reference for contextualizing model performance, it also highlights a limitation: some proximity between synthetic and real data is expected, which makes assessing memorization risk via Authenticity more challenging in small datasets. Another important consideration is that our privacy assessment relied exclusively on distance-based approaches. Since they primarily capture local similarity patterns, they might fail to detect subtler forms of memorization. Alternative privacy assessment techniques, such as membership inference attacks [44] or attribute inference attacks [45], could provide complementary insights and should be explored in future work.

Among all evaluated models, TVAE and Tab DDPM achieved the highest average performance scores, though their strengths differed across evaluation dimensions. TVAE excelled in several fidelity-related metrics, such as correlation preservation and variable-level similarity, while Tab DDPM performed best on utility-oriented metrics, including ML Efficiency and the Survival Metric. However, these gains came with important trade-offs. TVAE consistently exhibited concerning privacy behavior and produced high rates of domain constraint violations. Tab DDPM also demonstrated problematic privacy patterns on the ACTG dataset, whereas on AML, it achieved a potentially acceptable privacy-utility trade-off, performing just slightly below the test-to-train reference threshold. When maximizing utility is the primary goal, Tab DDPM may therefore represent an acceptable choice on AML, though caution remains warranted. Taking fidelity, utility, privacy, and clinical constraints into account, Survival CTGAN emerged as the most balanced model across both datasets. Finally, while we evaluated a diverse set of state-of-the-art generative models, large language model-based approaches, such as GReaT [46], were not included due to their extensive computational requirements for training, tuning, and sampling. Exploring these models in future work could provide additional insights.

While this study provides valuable insights, it has limitations. First, the analysis was limited to two datasets, ACTG (n = 1151) and AML (n = 1590), which are relatively small in size and may restrict the generalizability of our findings. However, these datasets were deliberately chosen because they represent high-quality clinical trial data with survival endpoints and pose a stringent challenge for generative models under realistic data-scarce conditions. Nevertheless, larger, multi-center cohorts will be required to validate and extend our recommendations on synthetic data generation in broader clinical settings. Second, while the chosen metrics emphasized utility and fidelity, they represent only a subset of the wide range of metrics available for evaluating synthetic data. Although this selection was guided by relevance to the study's goals, exploring additional metrics could provide a more comprehensive understanding. Third, the hyperparameter spaces of the generative models were predefined, which may have constrained the discovery of optimal configurations. Additionally, the study limited HPO to 30 optimization rounds, which may have restricted the ability to fully explore the optimization space, especially for models with large search spaces. Finally, while we assessed fundamental clinical validity constraints, future research should incorporate more complex, nuanced domain-specific constraints to better align synthetic datasets with real-world clinical scenarios. Addressing these limitations will facilitate the development of more reliable, generalizable, and clinically applicable synthetic datasets.

## 4. Conclusion

This study systematically evaluated four HPO objectives across nine generative models on two clinical trial datasets, aiming to determine the quality improvements achievable through HPO compared to default hyperparameters, and identifying optimal metrics to guide the optimization. Our experiments showed clear improvements through HPO, with Tab DDPM showing the largest relative gains (up to 335%) due to poor performance with default hyperparameters, followed by TVAE (up to 60%), CTGAN (up to 39%), and CTAB-GAN+ (up to 38%), strongly advocating for the computational investment in HPO. Compound metric optimization objectives consistently outperformed single-metric approaches, providing more balanced and broadly applicable synthetic datasets.

None of the evaluated generative models inherently learned to adhere to domain-specific constraints for survival data, highlighting the need for explicit validation steps. Despite better adherence to clinical constraints by survival-optimized models, these models did not universally outperform general-purpose models, underlining the importance of evaluating both approaches in clinical contexts. Pre- and postprocessing on the model and dataset level significantly improved constraint adherence, particularly for ensuring plausible OSTM and EFSTM relationships, which are critical for survival analysis.

Across all evaluated models, TVAE and Tab DDPM achieved the highest overall performance but showed important trade-offs: TVAE raised persistent privacy and constraint concerns, while Tab DDPM balanced utility and privacy better on AML but not on ACTG. Considering fidelity, utility, privacy, and clinical validity together, Survival CTGAN emerged as the most balanced model, highlighting the need for multi-dimensional evaluation.

Our analysis of evaluation metrics showed that high-variability metrics, such as the $S_{pMSE}$ Index, were more responsive to changes in data quality, while stable metrics like the Survival Metric offered consistency but limited sensitivity. Correlations among metrics revealed redundancies, underscoring the need to carefully balance their weighting in compound objectives.

Taken together, our findings suggest that systematic compound metric HPO, robust data preprocessing, explicit domain validation, careful metric selection, and dedicated privacy audits represent promising components for improving the reliability, clinical relevance, and overall quality of synthetic data generation workflows.

## CRediT authorship contribution statement

**Waldemar Hahn**: Conceptualization, Methodology, Writing – original draft, Visualization, Formal Analysis, Software.
**Jan-Niklas Eckardt**: Validation, Writing – review & editing.
**Christoph Röllig**: Data provision, Writing – review & editing.

**Martin Sedlmayr**: Writing – review & editing.
**Jan Moritz Middeke**: Writing – review & editing, Supervision.
**Markus Wolfien**: Methodology, Writing – review & editing, Validation, Supervision.

## Declaration of competing interest

The authors declare that they have no known competing financial interests or personal relationships that could have appeared to influence the work reported in this paper.

## Data availability

The AML dataset that has been used is confidential. The ACTG dataset is publicly available.

## Declaration of generative AI in scientific writing

During the preparation of this work the author(s) used ChatGPT 4o / 5 in order to correct the grammar, clarity, and conciseness of the text. After using this tool/service, the author(s) reviewed and edited the content as needed and take(s) full responsibility for the content of the published article.

## Tables

**Table 1**. Overview of AML and ACTG clinical trial datasets with patient counts and variable type distributions.

| Dataset | Patients | Total variables | Binary variables | Categorical variables | Integer variables | Float variables |
|---|---|---|---|---|---|---|
| AML | 1590 | 92 | 85 | 1 | 1 | 5 |
| ACTG | 1151 | 15 | 6 | 4 | 4 | 1 |

**Table 2**. Overview of generative models used.

| Model | Base Architecture | Survival Adaptation | Robust Preprocessing | Implementation |
|---|---|---|---|---|
| RTVAE [26] | GAN | No | Yes | Synthcity |
| TVAE [23] | VAE | No | No | Original |
| CTGAN [23] | GAN | No | No | Original |
| CTAB-GAN+ [25] | GAN | No | Yes | Original |
| Tab DDPM [12] | DDPM | No | Yes | Original |
| SURVAE | VAE (TVAE) | Yes | Yes | Synthcity |
| Survival GAN [27] | GAN | Yes | Yes | Synthcity |
| Survival CTGAN | GAN (CTGAN) | Yes | Yes | Synthcity |
| Survival NFlow | NFlow | Yes | Yes | Synthcity |

**Table 3**. Summary of evaluation metrics with objectives and key methodological details.

| Metric | Objective | Key Details |
|---|---|---|
| Basic Statistical Measure [28] | Assess numerical distribution similarity | Compares means, medians, and standard deviations; computes and averages relative errors across numerical variables |
| Regularized Support Coverage [28] | Evaluate reproduction of variable support | Measures the proportion of the real data's variable support captured by the synthetic data; numerical variables are binned into 10 intervals |
| Log-transformed Correlation Score [28] | Assess preservation of pairwise correlations | Uses Pearson's (continuous pairs), correlation ratio (continuous–categorical), and Theil's U (categorical pairs) with a log transformation to moderate small differences |
| $S_{pMSE}$ Index [28] | Distinguish real vs. synthetic data | Compares observed pMSE to expected pMSE ($pMSE_0$) and normalizes the ratio using an alpha of 1.2 |
| ML Efficiency | Evaluate predictive utility on real data | Uses CatBoost (optimized on real data), MCC metric to measure absolute predictive performance, independent of baseline characteristics |
| K-Means Score | Assess overall dataset-level similarity | Runs k-means (with k=10) on real data to fix centroids; synthetic data are clustered using these centroids and relative frequencies are compared (with each cluster capped at 1) |
| Survival Metric [27] | Evaluate similarity in survival outcomes | Averages three KM-based metrics (Optimism, Short-sightedness, Kaplan–Meier Divergence), each rescaled to [0, 1], to assess the match between synthetic and real survival curves |

**Table 4**. Comparison of HPO objectives across nine generative models on the ACTG and AML datasets. Each entry shows the mean ± standard deviation over 25 synthetic datasets per model-objective pair (5 training seeds × 5 sampling seeds). To determine whether each HPO objective performance differs significantly from the default hyperparameters, the five sampling seed scores were averaged within each training seed, yielding n = 5 independent values per model-objective. An omnibus Kruskal-Wallis test evaluated overall differences across objectives, followed by two-sided Mann-Whitney U tests versus the default hyperparameters with Holm correction for multiple comparisons. * indicates Holm-adjusted $p < 0.05$.

| Dataset | Model \ HPO objective | Default | Survival Objective | ML Objective | Four Metrics Objective | Full Objective |
|---|---|---|---|---|---|---|
| ACTG | RTVAE | 0.6415 ± 0.0221 | 0.6474 ± 0.0099 | 0.6522 ± 0.0169 | 0.6779 ± 0.0191 | 0.6856* ± 0.0138 |
| | TVAE | 0.5058 ± 0.0130 | 0.7910* ± 0.0151 | 0.7708* ± 0.0152 | 0.7904* ± 0.0215 | 0.7740* ± 0.0156 |
| | CTGAN | 0.6053 ± 0.0376 | 0.5840 ± 0.0407 | 0.6525 ± 0.0171 | 0.7347* ± 0.0121 | 0.7308* ± 0.0196 |
| | CTAB-GAN+ | 0.6049 ± 0.0408 | 0.7254* ± 0.0277 | 0.6125 ± 0.0246 | 0.7489* ± 0.0365 | 0.7559* ± 0.0324 |
| | Tab DDPM | 0.5710 ± 0.0319 | 0.6505* ± 0.0313 | 0.3601* ± 0.0155 | 0.7895* ± 0.0136 | 0.7881* ± 0.0187 |
| | SURVAE | 0.6848 ± 0.0181 | 0.7170* ± 0.0213 | 0.6714 ± 0.0152 | 0.7244* ± 0.0143 | 0.7437* ± 0.0145 |
| | SURVIVAL GAN | 0.6591 ± 0.0352 | 0.6356 ± 0.0250 | 0.6535 ± 0.0211 | 0.6451 ± 0.0122 | 0.6556 ± 0.0266 |
| | SURVIVAL CTGAN | 0.6716 ± 0.0272 | 0.7440* ± 0.0184 | 0.5918* ± 0.0329 | 0.7589* ± 0.0142 | 0.7625* ± 0.0151 |
| | SURVIVAL NFlow | 0.6501 ± 0.0319 | 0.7063* ± 0.0196 | 0.7443* ± 0.0145 | 0.7182* ± 0.0173 | 0.7148* ± 0.0327 |
| | **Average** | 0.6216 | 0.6890 | 0.6343 | 0.7320 | **0.7346** |
| AML | RTVAE | 0.5575 ± 0.0658 | 0.5640 ± 0.1072 | 0.5692 ± 0.0585 | 0.6003 ± 0.1583 | 0.5701 ± 0.0632 |
| | TVAE | 0.4747 ± 0.0076 | 0.5422* ± 0.0095 | 0.7189* ± 0.0117 | 0.7538* ± 0.0090 | 0.7611* ± 0.0101 |
| | CTGAN | 0.5073 ± 0.0480 | 0.5787 ± 0.0171 | 0.6020 ± 0.0221 | 0.7008* ± 0.0121 | 0.6876* ± 0.0104 |
| | CTAB-GAN+ | 0.5376 ± 0.0388 | 0.6508* ± 0.0628 | 0.7475* ± 0.0523 | 0.6524* ± 0.0570 | 0.6641* ± 0.0612 |
| | Tab DDPM | 0.2280 ± 0.0280 | 0.2583 ± 0.0061 | 0.6473* ± 0.0310 | 0.7281* ± 0.0095 | 0.7658* ± 0.0093 |
| | SURVAE | 0.6151 ± 0.0147 | 0.6587 ± 0.0400 | 0.5730* ± 0.0354 | 0.6966 ± 0.0432 | 0.6842 ± 0.0530 |
| | SURVIVAL GAN | 0.6649 ± 0.0145 | 0.7238* ± 0.0141 | 0.7321* ± 0.0093 | 0.7291* ± 0.0112 | 0.7124* ± 0.0115 |
| | SURVIVAL CTGAN | 0.6662 ± 0.0426 | 0.6690 ± 0.0175 | 0.7408* ± 0.0089 | 0.7341* ± 0.0124 | 0.7505* ± 0.0083 |
| | SURVIVAL NFlow | 0.5722 ± 0.0191 | 0.6982* ± 0.0141 | 0.7046* ± 0.0169 | 0.7176* ± 0.0120 | 0.7139* ± 0.0131 |
| | **Average** | 0.5359 | 0.5937 | 0.6706 | **0.7014** | 0.7011 |

**Table 5.** HPO results across different optimization goals. The upper half indicates, for each optimization goal, the number of models (n = 9) on which each HPO objective achieved the best performance. The lower half shows mean ± standard error (SE) across models (n = 9) for each optimization target. For each model-objective pair, the score is the average over 25 synthetic datasets (5 training seeds × 5 sampling seeds). To assess whether HPO objectives significantly differ from the default hyperparameters, an omnibus Friedman test (blocking by model) was applied. When significant, two-sided Wilcoxon signed-rank tests pairing by model compared each strategy to the default, with Holm correction for multiple comparisons. * indicates Holm-adjusted p < 0.05.

| Dataset | Task \ HPO objective | Default | Survival Objective | ML Objective | Four Metrics Objective | Full Objective |
|---|---|---|---|---|---|---|
| ACTG | ML Efficiency (# best) | 0 | 1 | 1 | 3 | **4** |
| | Survival Metric (# best) | 0 | 2 | 0 | **6** | 1 |
| | Average of four metrics (# best) | 0 | 0 | 1 | 3 | **5** |
| | Average of all metrics (# best) | 1 | 1 | 1 | 2 | **4** |
| AML | ML Efficiency (# best) | 0 | 2 | **4** | 2 | 1 |
| | Survival Metric (# best) | 0 | **5** | 2 | 1 | 1 |
| | Average of four metrics (# best) | 0 | 0 | 2 | **4** | 3 |
| | Average of all metrics (# best) | 0 | 0 | 2 | **4** | 3 |
| | **Average values HPO objectives** | | | | | |
| ACTG | ML Efficiency (average) | 0.0444 ± 0.0081 | 0.0439 ± 0.0109 | 0.0387 ± 0.0106 | 0.0529 ± 0.0135 | **0.0718** ± 0.0116 |
| | Survival Metric (average) | 0.9744 ± 0.0065 | 0.9829* ± 0.0051 | 0.9748 ± 0.0076 | **0.9841*** ± 0.0053 | 0.9827* ± 0.0051 |
| | Average of four metrics (average) | 0.4946 ± 0.0236 | 0.5747* ± 0.0245 | 0.5248 ± 0.0311 | **0.6235*** ± 0.0168 | 0.6233* ± 0.0156 |
| | Average of all metrics (average) | 0.6216 ± 0.0189 | 0.6890 ± 0.0214 | 0.6343 ± 0.0392 | 0.7320* ± 0.0160 | **0.7346*** ± 0.0143 |
| AML | ML Efficiency (average) | 0.1222 ± 0.0433 | 0.1809 ± 0.0353 | 0.2556 ± 0.0149 | **0.2571*** ± 0.0165 | 0.2410 ± 0.0180 |
| | Survival Metric (average) | 0.8723 ± 0.0250 | 0.9303* ± 0.0117 | **0.9341*** ± 0.0141 | 0.9284* ± 0.0177 | 0.9278* ± 0.0170 |
| | Average of four metrics (average) | 0.5008 ± 0.0433 | 0.5603* ± 0.0453 | 0.6403* ± 0.0261 | **0.6799*** ± 0.0160 | 0.6711* ± 0.0234 |
| | Average of all metrics (average) | 0.5359 ± 0.0443 | 0.5937* ± 0.0467 | 0.6706* ± 0.0245 | **0.7014*** ± 0.0159 | 0.7011* ± 0.0202 |

**Table 6**. Impact of non-valid patient removal on evaluation metrics. The table shows the number of models that improved after patient removal for each metric on the ACTG and AML datasets, along with the average metric differences (after - before) across nine models. Significance was assessed with a two-sided Wilcoxon signed-rank test (paired by model) after averaging all values to obtain one value per model. P-values were Holm-adjusted across metrics within each dataset. * indicates adjusted $p < 0.05$.

| Metric | # better on ACTG | # better on AML | # better total | avg dif ACTG | avg dif AML |
|---|---|---|---|---|---|
| Basic Statistical Measure | 1/9 | 1/9 | 2/18 | -0.0032 | -0.0223 |
| Log-trans Correlation Score | 3/9 | 3/9 | 6/18 | -0.0047 | -0.0164 |
| Regularized Support Coverage | 0/9 | 1/9 | 1/18 | -0.0176* | -0.0087 |
| K-Means Score | 1/9 | 1/9 | 2/18 | -0.0037 | -0.0123 |
| ML Efficiency MCC EFS | 4/9 | - | - | -0.0119 | - |
| ML Efficiency MCC OS | - | 1/9 | - | - | -0.0081 |
| ML Efficiency MCC CR1 | - | 0/9 | - | - | -0.0228* |
| Survival Metric | 4/9 | 5/9 | 9/18 | 0.0014 | -0.0038 |
| $S_{pMSE}$ Index | 4/9 | 7/9 | 11/18 | 0.0033 | 0.0225 |
| **average** | 1/9 | 1/9 | 2/18 | -0.0052 | -0.0090 |

## Figures

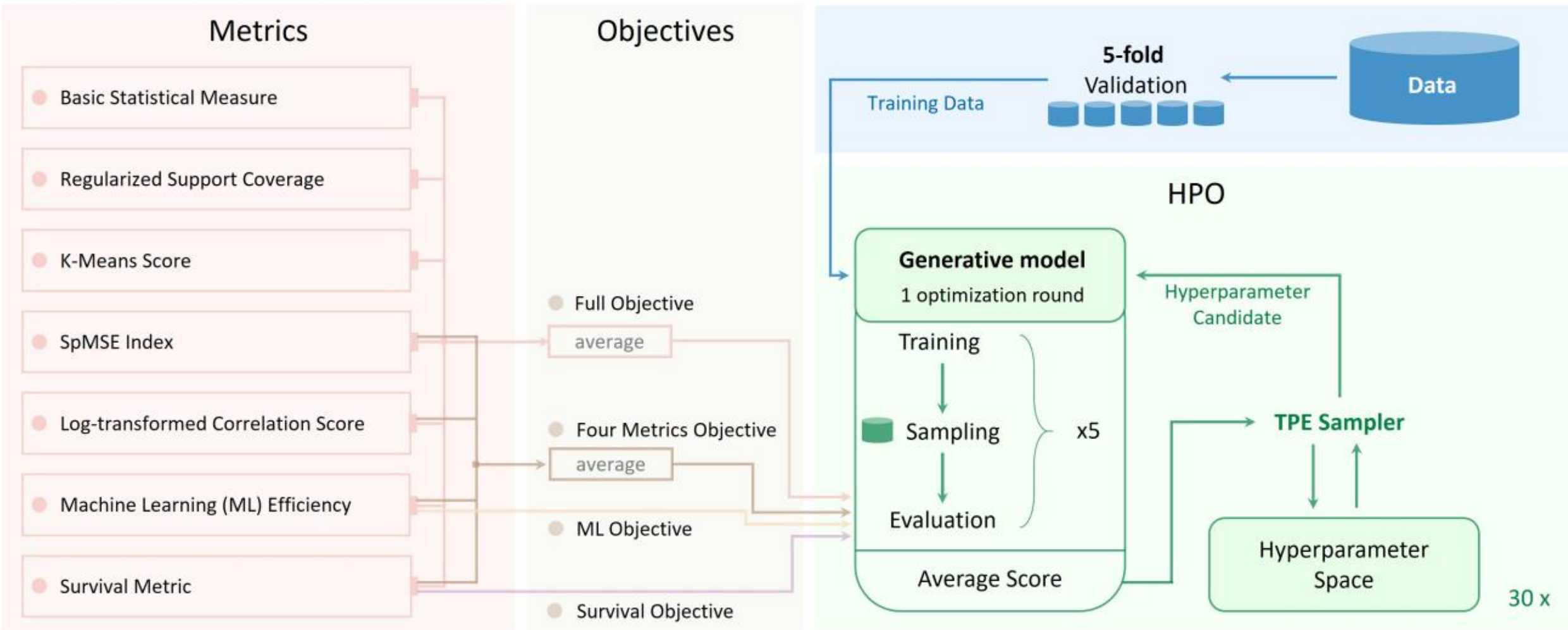

**Figure 1**. Overview of the HPO process. The left side illustrates the evaluation metrics and how they are combined into four different optimization objectives. The right side depicts the HPO workflow using a Tree-structured Parzen Estimator (TPE) Sampler. Each trial consists of five rounds, where the generative model is trained on four of the five cross-validation folds and evaluated according to the selected optimization objective. The trial score is computed as the average across these five rounds. After 30 trials, the best-performing hyperparameter configuration for each objective is saved. This process is repeated for all nine generative models.

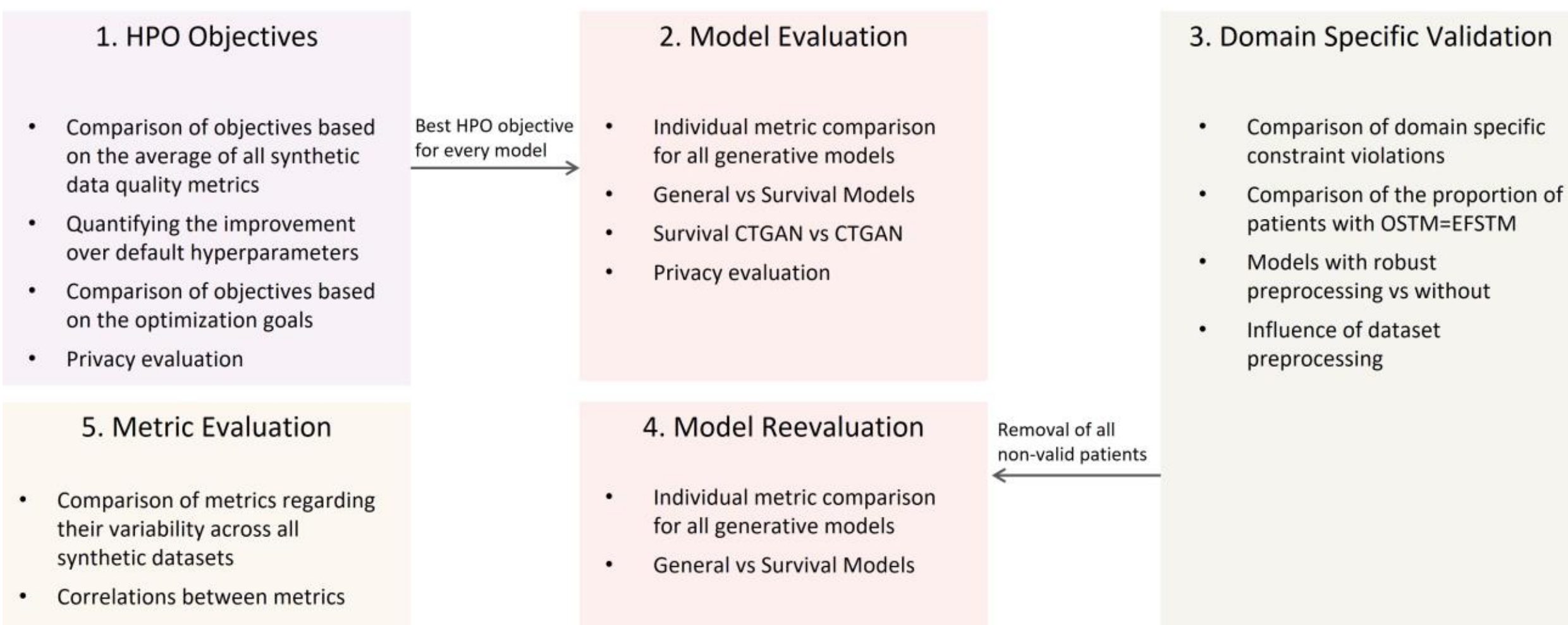

**Figure 2**. Overview of the evaluation framework. The process is divided into five main components: (1) Comparison of the hyperparameter optimization objectives; (2) model evaluation, including a comparison of general models and survival-optimized models; (3) domain-specific validation, focusing on constraint violations and preprocessing influences; (4) model reevaluation after removing invalid data; (5) metric evaluation, examining metric variability and inter-metric correlations.

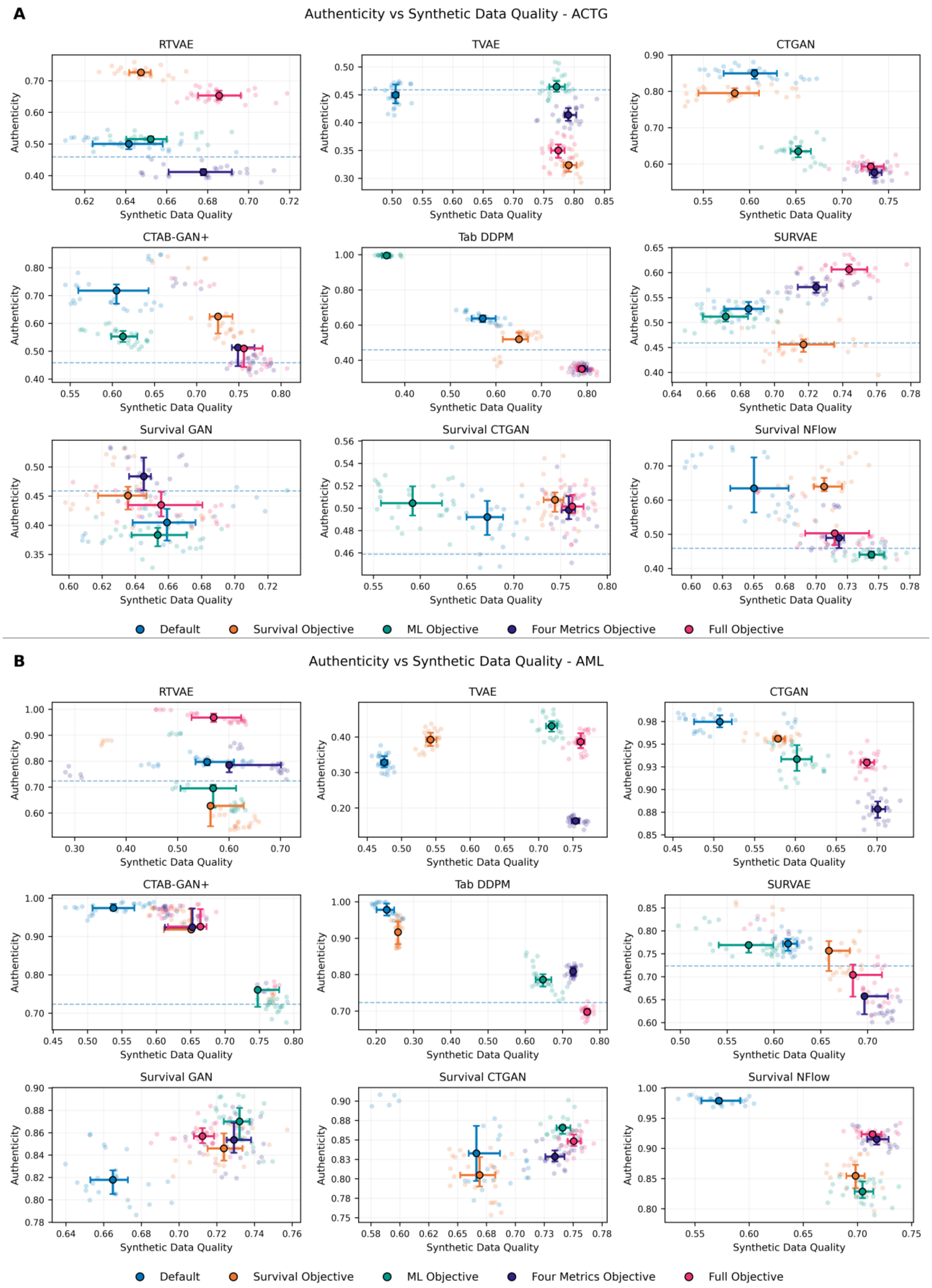
A
Authenticity vs Synthetic Data Quality - ACTG
RTVAE
TVAE
CTGAN
CTAB-GAN+
Tab DDPM
SURVAE
Survival GAN
Survival CTGAN
Survival NFlow
Authenticity
Synthetic Data Quality
Default
Survival Objective
ML Objective
Four Metrics Objective
Full Objective
B
Authenticity vs Synthetic Data Quality - AML
RTVAE
TVAE
CTGAN
CTAB-GAN+
Tab DDPM
SURVAE
Survival GAN
Survival CTGAN
Survival NFlow
Authenticity
Synthetic Data Quality
Default
Survival Objective
ML Objective
Four Metrics Objective
Full Objective

**Figure 3**. Comparison of HPO objectives regarding Authenticity vs. Synthetic Data Quality for each generative model on (A) ACTG and (B) AML datasets. Each point represents one synthetic dataset generated under a specific hyperparameter configuration. The mean per configuration is shown as a black-edged circle, with whiskers indicating the interquartile range (IQR) for Synthetic Data Quality (horizontal) and Authenticity (vertical). The dashed blue line marks the real-data reference, obtained by treating the real test set as synthetic and computing its Authenticity score against the training set. Configurations with Authenticity values below this reference indicate an increased memorization risk and indicate higher privacy risk.

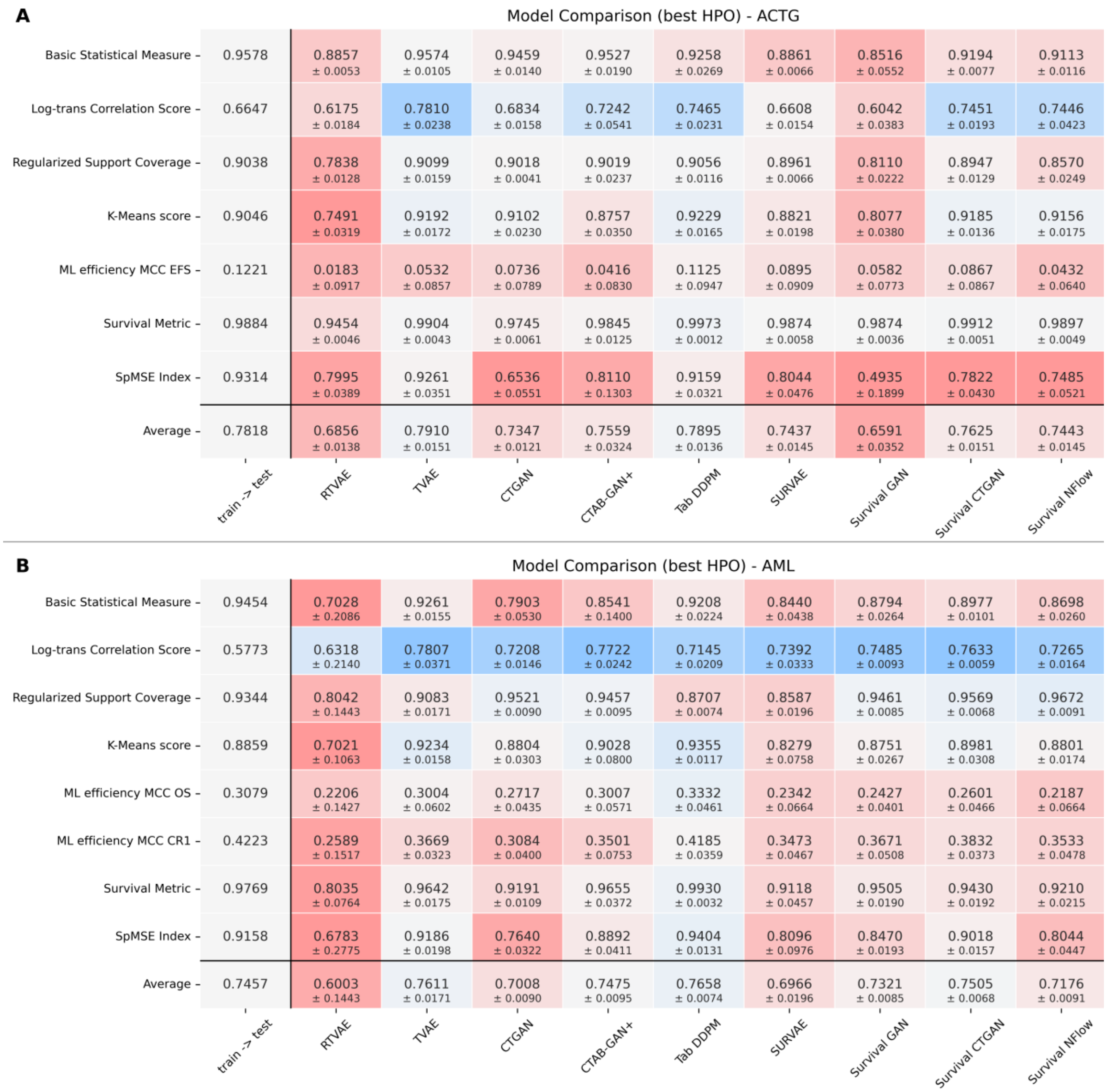

**Figure 4**. Comparison of generative models using their respective best HPO objective. The heatmaps show the average metric scores ± standard deviation across 25 synthetic datasets for each model on (A) the ACTG dataset and (B) the AML dataset. For reference, the same evaluation metrics were applied to real data, treating the training set as if it were synthetic and using the test set as the ground truth. Color intensity reflects deviation from the real-data reference: values closer to the real baseline are shown in neutral tones, higher values are shaded blue, and lower values are shaded red.

**A**

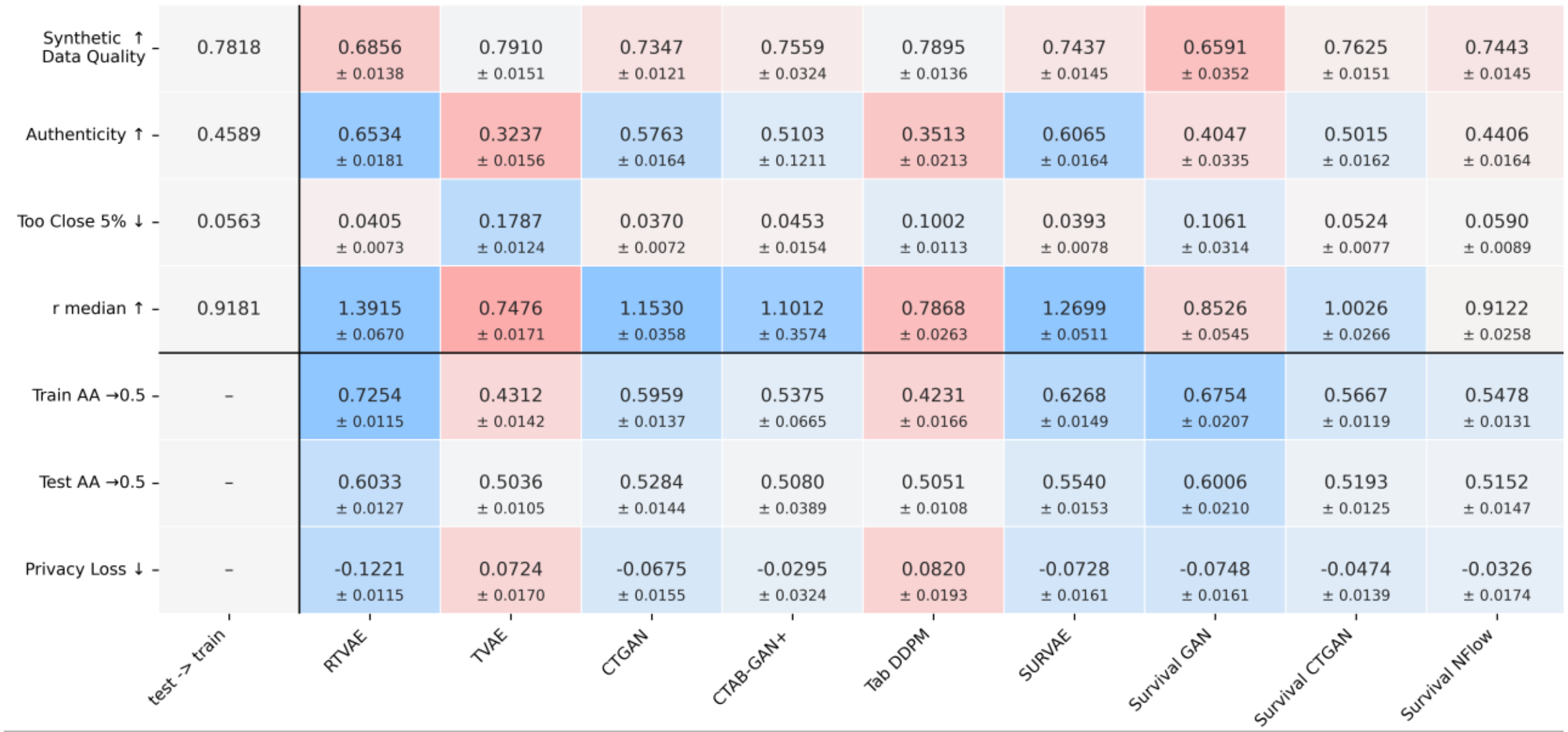

Privacy Evaluation (best HPO) - ACTG

| | test -> train | RTVAE | TVAE | CTGAN | CTAB-GAN+ | Tab DDPM | SURVAE | Survival GAN | Survival CTGAN | Survival NFlow |
|---|---|---|---|---|---|---|---|---|---|---|
| Synthetic Data Quality ↑ | 0.7818 | 0.6856 ± 0.0138 | 0.7910 ± 0.0151 | 0.7347 ± 0.0121 | 0.7559 ± 0.0324 | 0.7895 ± 0.0136 | 0.7437 ± 0.0145 | 0.6591 ± 0.0352 | 0.7625 ± 0.0151 | 0.7443 ± 0.0145 |
| Authenticity ↑ | 0.4589 | 0.6534 ± 0.0181 | 0.3237 ± 0.0156 | 0.5763 ± 0.0164 | 0.5103 ± 0.1211 | 0.3513 ± 0.0213 | 0.6065 ± 0.0164 | 0.4047 ± 0.0335 | 0.5015 ± 0.0162 | 0.4406 ± 0.0164 |
| Too Close 5% ↓ | 0.0563 | 0.0405 ± 0.0073 | 0.1787 ± 0.0124 | 0.0370 ± 0.0072 | 0.0453 ± 0.0154 | 0.1002 ± 0.0113 | 0.0393 ± 0.0078 | 0.1061 ± 0.0314 | 0.0524 ± 0.0077 | 0.0590 ± 0.0089 |
| r median ↑ | 0.9181 | 1.3915 ± 0.0670 | 0.7476 ± 0.0171 | 1.1530 ± 0.0358 | 1.1012 ± 0.3574 | 0.7868 ± 0.0263 | 1.2699 ± 0.0511 | 0.8526 ± 0.0545 | 1.0026 ± 0.0266 | 0.9122 ± 0.0258 |
| Train AA →0.5 | - | 0.7254 ± 0.0115 | 0.4312 ± 0.0142 | 0.5959 ± 0.0137 | 0.5375 ± 0.0665 | 0.4231 ± 0.0166 | 0.6268 ± 0.0149 | 0.6754 ± 0.0207 | 0.5667 ± 0.0119 | 0.5478 ± 0.0131 |
| Test AA →0.5 | - | 0.6033 ± 0.0127 | 0.5036 ± 0.0105 | 0.5284 ± 0.0144 | 0.5080 ± 0.0389 | 0.5051 ± 0.0108 | 0.5540 ± 0.0153 | 0.6006 ± 0.0210 | 0.5193 ± 0.0125 | 0.5152 ± 0.0147 |
| Privacy Loss ↓ | - | -0.1221 ± 0.0115 | 0.0724 ± 0.0170 | -0.0675 ± 0.0155 | -0.0295 ± 0.0324 | 0.0820 ± 0.0193 | -0.0728 ± 0.0161 | -0.0748 ± 0.0161 | -0.0474 ± 0.0139 | -0.0326 ± 0.0174 |

**B**

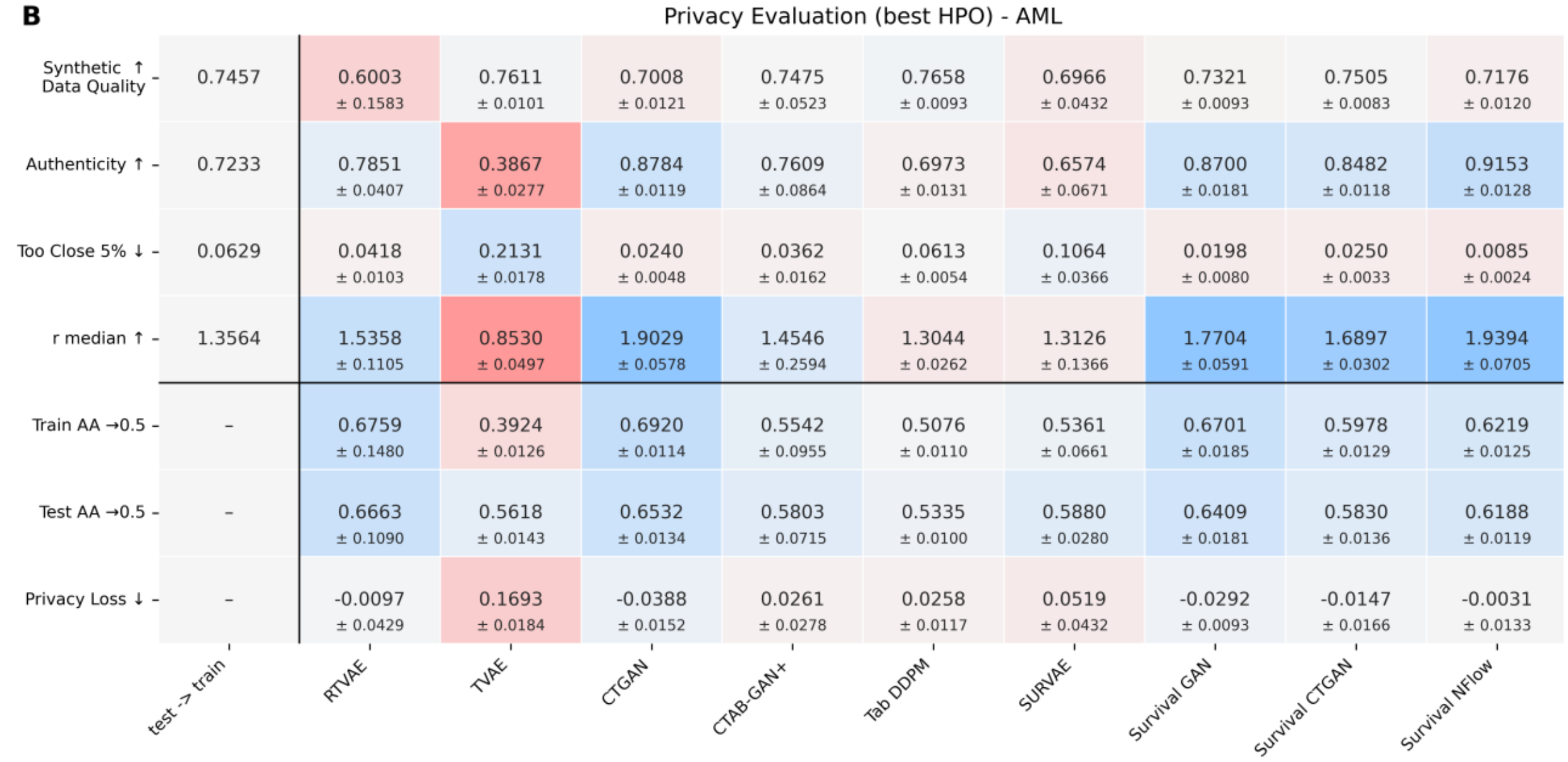

Privacy Evaluation (best HPO) - AML

| | test -> train | RTVAE | TVAE | CTGAN | CTAB-GAN+ | Tab DDPM | SURVAE | Survival GAN | Survival CTGAN | Survival NFlow |
|---|---|---|---|---|---|---|---|---|---|---|
| Synthetic Data Quality ↑ | 0.7457 | 0.6003 ± 0.1583 | 0.7611 ± 0.0101 | 0.7008 ± 0.0121 | 0.7475 ± 0.0523 | 0.7658 ± 0.0093 | 0.6966 ± 0.0432 | 0.7321 ± 0.0093 | 0.7505 ± 0.0083 | 0.7176 ± 0.0120 |
| Authenticity ↑ | 0.7233 | 0.7851 ± 0.0407 | 0.3867 ± 0.0277 | 0.8784 ± 0.0119 | 0.7609 ± 0.0864 | 0.6973 ± 0.0131 | 0.6574 ± 0.0671 | 0.8700 ± 0.0181 | 0.8482 ± 0.0118 | 0.9153 ± 0.0128 |
| Too Close 5% ↓ | 0.0629 | 0.0418 ± 0.0103 | 0.2131 ± 0.0178 | 0.0240 ± 0.0048 | 0.0362 ± 0.0162 | 0.0613 ± 0.0054 | 0.1064 ± 0.0366 | 0.0198 ± 0.0080 | 0.0250 ± 0.0033 | 0.0085 ± 0.0024 |
| r median ↑ | 1.3564 | 1.5358 ± 0.1105 | 0.8530 ± 0.0497 | 1.9029 ± 0.0578 | 1.4546 ± 0.2594 | 1.3044 ± 0.0262 | 1.3126 ± 0.1366 | 1.7704 ± 0.0591 | 1.6897 ± 0.0302 | 1.9394 ± 0.0705 |
| Train AA →0.5 | - | 0.6759 ± 0.1480 | 0.3924 ± 0.0126 | 0.6920 ± 0.0114 | 0.5542 ± 0.0955 | 0.5076 ± 0.0110 | 0.5361 ± 0.0661 | 0.6701 ± 0.0185 | 0.5978 ± 0.0129 | 0.6219 ± 0.0125 |
| Test AA →0.5 | - | 0.6663 ± 0.1090 | 0.5618 ± 0.0143 | 0.6532 ± 0.0134 | 0.5803 ± 0.0715 | 0.5335 ± 0.0100 | 0.5880 ± 0.0280 | 0.6409 ± 0.0181 | 0.5830 ± 0.0136 | 0.6188 ± 0.0119 |
| Privacy Loss ↓ | - | -0.0097 ± 0.0429 | 0.1693 ± 0.0184 | -0.0388 ± 0.0152 | 0.0261 ± 0.0278 | 0.0258 ± 0.0117 | 0.0519 ± 0.0432 | -0.0292 ± 0.0093 | -0.0147 ± 0.0166 | -0.0031 ± 0.0133 |

**Figure 5**. Privacy Evaluation of generative models with their best hyperparameter configurations. Each entry reports the mean ± standard deviation across 25 synthetic datasets for each model on (A) the ACTG dataset and (B) the AML dataset. Synthetic Data Quality represents the average of all fidelity and utility metrics, while the remaining metrics assess privacy. Arrows indicate the desired direction for each metric. A real (test-to-train) reference for Authenticity and related summaries is provided. Results falling below this reference suggest increased memorization risk and indicate privacy concerns. Color intensity reflects deviation from the real-data reference: values closer to the baseline are shown in neutral tones, higher values are shaded blue, and lower values are shaded red. For Train AA and Test AA, values close to 0.5 are shown in neutral colors, higher values in blue, and lower (worse) values in red. Positive Privacy Loss is displayed in red, while negative values are shown in blue.

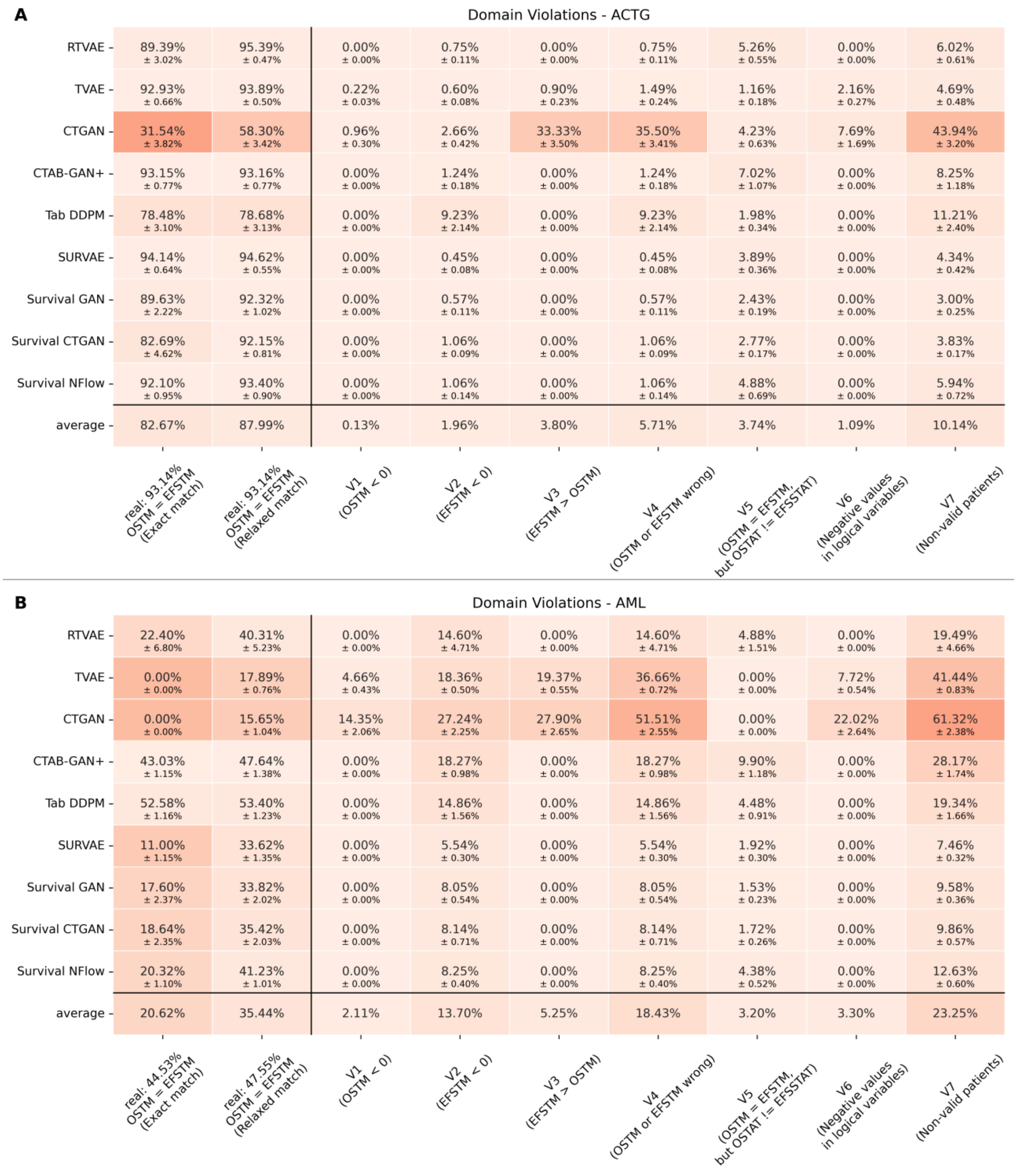

**Figure 6.** Domain violations in synthetic datasets for (A) ACTG and (B) AML, evaluating adherence to logical constraints (*V1–V7*). Each heatmap cell shows the mean ± standard error (SE) percentage of patients violating a constraint for a given model. SEs are based on 25 cluster means per model (5 hyperparameter configurations × 5 training seeds), where each cluster mean first averages 5 sampling seeds. Differences in violation rates reflect the impact of preprocessing and model design. The proportion of patients with matching Overall Survival Time (OSTM) and Event-Free Survival Time (EFSTM) are also reported, providing insight into the models' ability to maintain key survival relationships.

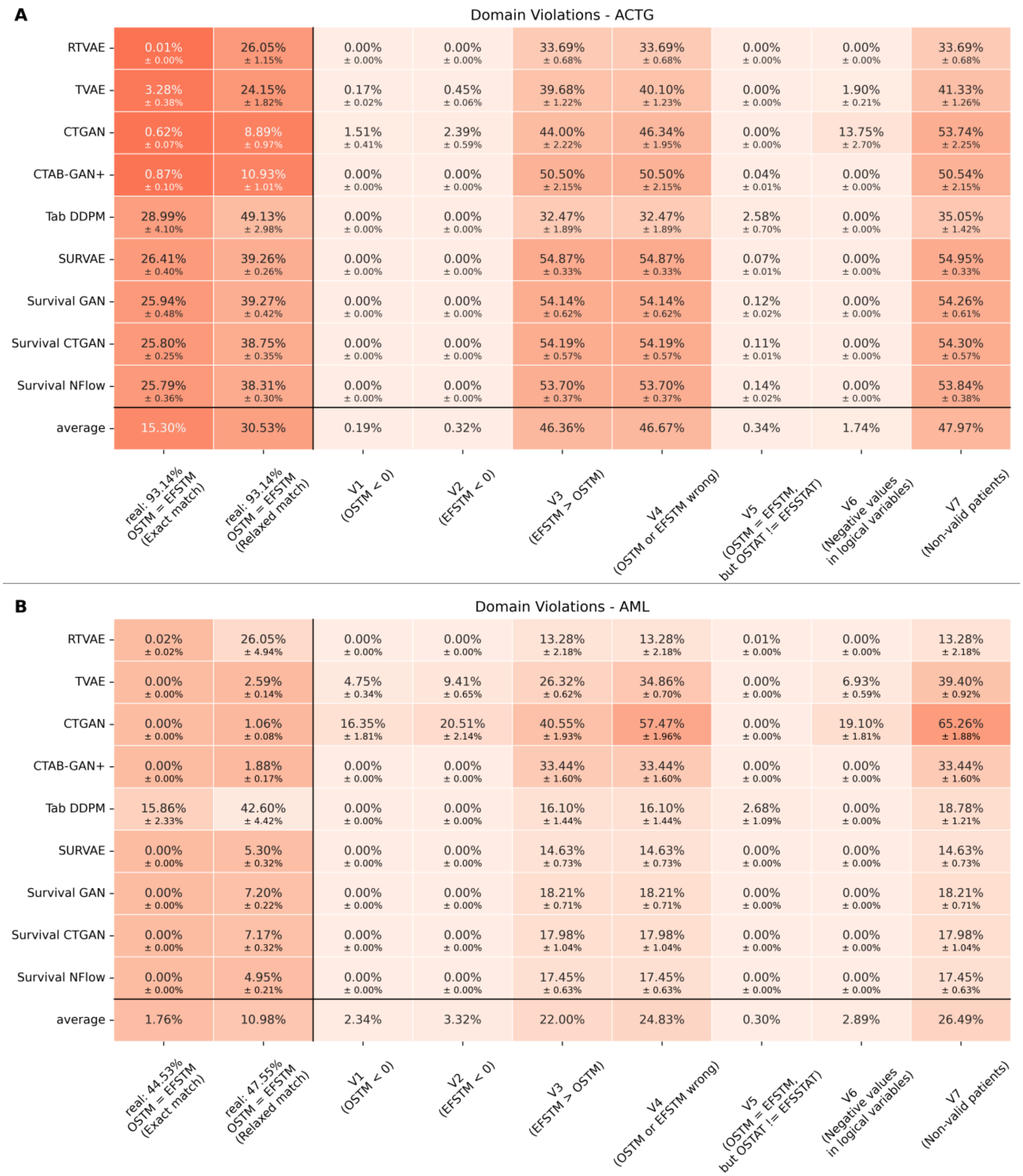

**Figure 7.** Domain violations in synthetic datasets for (A) ACTG and (B) AML after removing the EFSTM transformation. Each heatmap cell reports the mean ± standard error (SE) percentage of patients violating a constraint (*V1*–*V7*) for a given model. SEs are computed from 25 cluster means per model (5 hyperparameter configurations × 5 training seeds), where each cluster mean first averages 5 sampling seeds. Removing the EFSTM transformation increased fault rates, particularly for EFSTM exceeding OSTM (*V3*), demonstrating the impact of preprocessing on maintaining logical consistency. The proportion of patients with matching OSTM and EFSTM times also decreased, highlighting the role of preprocessing in preserving key survival relationships.

# Appendix

## Tables (Appendix)

**Table A1**. Hyperparameter spaces used for HPO of TVAE, CTGAN, CTAB-GAN+, and Tab DDPM. The remaining five generative models utilized predefined hyperparameter spaces from the Synthcity framework (https://github.com/vanderschaarlab/synthcity).

| Parameter \ Model | TVAE | CTGAN | CTAB-GAN+ | Tab DDPM |
|---|---|---|---|---|
| Learning rate (lr) | 0.00002 – 0.002 (log scale) | *generator_lr* and *discriminator_lr*: 0.00002 – 0.002 (log scale) | 0.00002 – 0.002 (log scale) | 0.00001 – 0.003 (log scale) |
| Epochs | 300, 500, 1000, 5000, 10000 | 100, 300, 500, 1000, 5000 | 100, 300, 500, 1000, 5000 | 500, 1000, 2500, 5000, 7500, 10000 |
| Layer Count | 1, 2, 3, 4 | 1, 2, 3, 4 (for generator and discriminator) | 1, 2, 3, 4 | 1, 2, 3, 4 (for MLP) |
| First Layer Dimension | 64, 128, 256, 512 | 64, 128, 256, 512 | 64, 128, 256 | 128, 256, 512, 1024 (for MLP) |
| Middle Layer Dimension | 64, 128, 256, 512 (*must decrease for compression network; decompression is the reverse order*) | 64, 128, 256, 512 (*fixed for all middle layers*) | 64, 128, 256 (*fixed for all middle layers*) | 128, 256, 512, 1024 (for MLP, fixed for all middle layers) |
| Last Layer Dimension | 64, 128, 256, 512 | 64, 128, 256, 512 | 64, 128, 256 | 128, 256, 512, 1024 (for MLP) |
| Batch Size | 20, 50, 100, 200, 500, 1000 | 20, 50, 100, 200, 500, 1000 | 128, 256, 512, 1024 | 64, 128, 256, 512, 1024 |
| Random Dimension | - | - | 16, 32, 64, 128 | |
| Number of Channels | - | - | 16, 32, 64 | |
| Embedding Dimension | 16, 32, 64, 128, 256 | 16, 32, 64, 128, 256 | - | |
| Loss Factor | 0.001 – 10 (log scale) | - | - | |
| Log Frequency | - | True, False | - | |
| Number of Timesteps | - | - | - | 100, 250, 500, 750, 1000 |
| Weight Decay | - | - | - | 0.0, 1e-5, 1e-4, 1e-3 |

**Table A2**. HPO durations (in elapsed hours) for each optimization objective, accumulated across all nine generative models on ACTG and AML datasets.

| | Survival Objective | ML Objective | Four Metrics Objective | Full Objective |
|---|---|---|---|---|
| Optimization duration ACTG (in elapsed hours) | 49.38 | 18.70 | 60.28 | 58.31 |
| Optimization duration AML (in elapsed hours) | 127.19 | 172.75 | 142.89 | 232.34 |
| **Total** | 176.57 | 191.45 | 203.17 | 290.65 |

**Table A3**. Total HPO durations (in elapsed hours) for each generative model on ACTG and AML datasets, accumulated across all four HPO objectives.

| | **RTVAE** | **TVAE** | **CTGAN** | **CTAB-GAN+** | **Tab DDPM** | **SURVAE** | **Survival GAN** | **Survival CTGAN** | **Survival NFlow** | **Total** |
|---|---|---|---|---|---|---|---|---|---|---|
| Optimization duration ACTG (in elapsed hours) | 3.51 | 47.42 | 43.03 | 18.81 | 10.50 | 4.42 | 11.48 | 15.75 | 31.75 | 186.67 |
| Optimization duration AML (in elapsed hours) | 14.92 | 230.44 | 120.96 | 130.15 | 41.64 | 15.11 | 25.53 | 61.65 | 34.78 | 675.17 |
| **Total** | 18.43 | 277.86 | 163.99 | 148.96 | 52.14 | 19.53 | 37.01 | 77.40 | 66.53 | 861.84 |

Figures (Appendix)

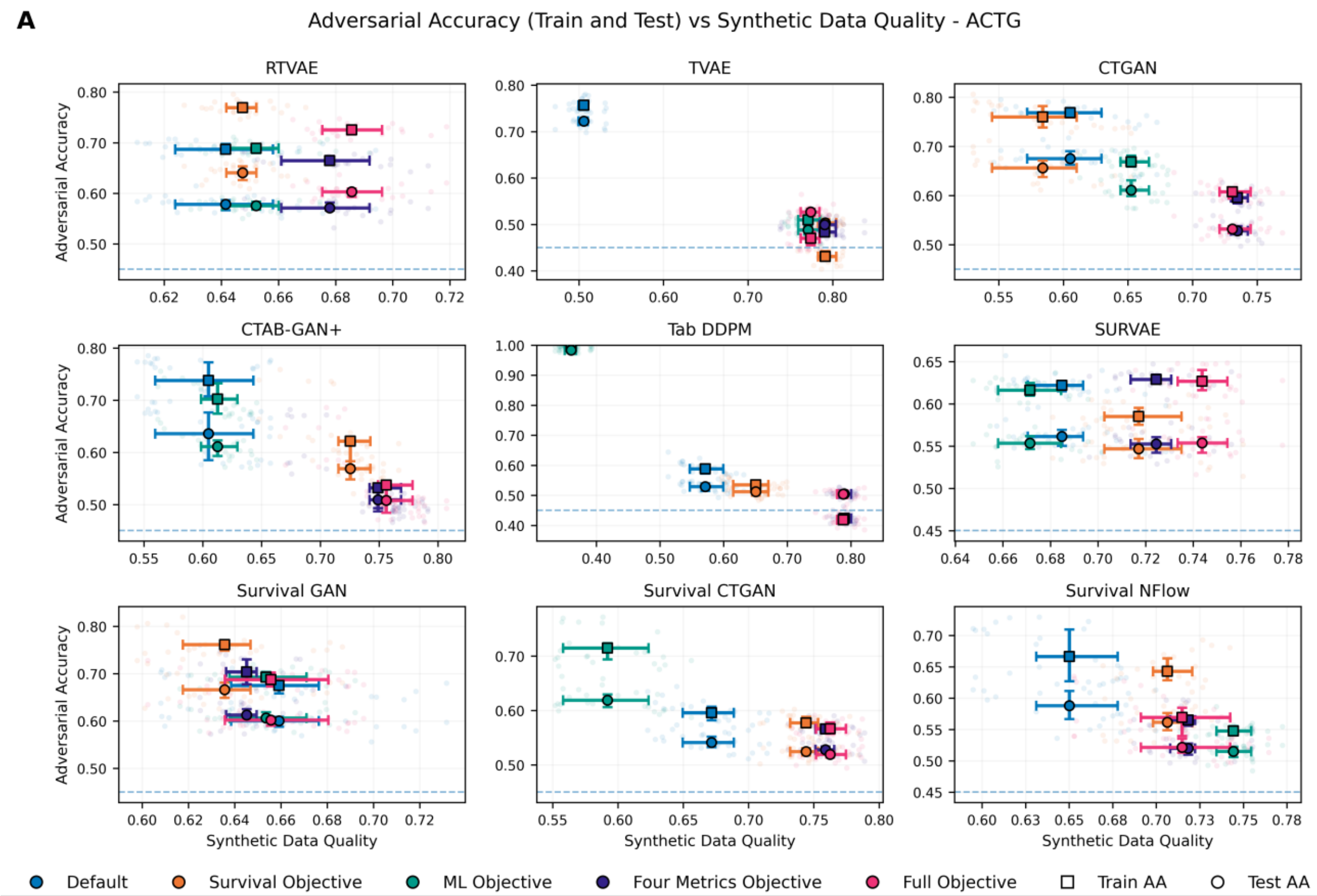
A
Adversarial Accuracy (Train and Test) vs Synthetic Data Quality - ACTG
RTVAE
TVAE
CTGAN
CTAB-GAN+
Tab DDPM
SURVAE
Survival GAN
Survival CTGAN
Survival NFlow
Adversarial Accuracy
Synthetic Data Quality
Default
Survival Objective
ML Objective
Four Metrics Objective
Full Objective
Train AA
Test AA

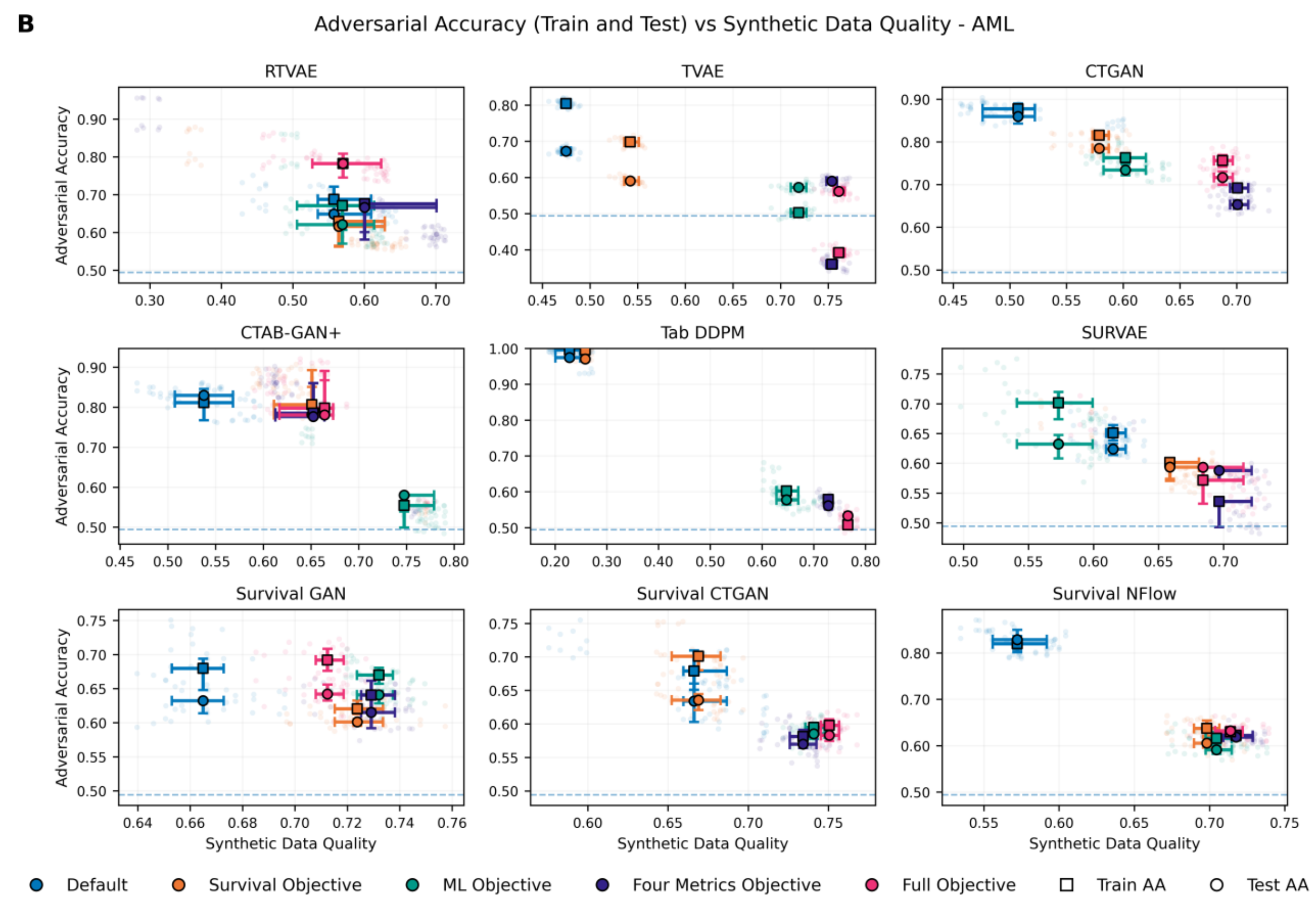
B
Adversarial Accuracy (Train and Test) vs Synthetic Data Quality - AML
RTVAE
TVAE
CTGAN
CTAB-GAN+
Tab DDPM
SURVAE
Survival GAN
Survival CTGAN
Survival NFlow
Adversarial Accuracy
Synthetic Data Quality
Default
Survival Objective
ML Objective
Four Metrics Objective
Full Objective
Train AA
Test AA

**Figure A1**. Comparison of HPO objective results regarding Adversarial Accuracy (AA) vs. Synthetic Data Quality for each generative model on (A) ACTG and (B) AML datasets. Each point represents one synthetic dataset generated under a specific hyperparameter configuration. The square markers indicate Train AA, and the circular markers represent Test AA, with the mean per configuration outlined in black. Whiskers show the interquartile range (IQR) horizontally for Synthetic Data Quality and vertically for AA. The dashed blue line represents the real-data reference obtained by computing AA between the real training and test sets (expected ≈ 0.5). Configurations with Train AA substantially below this reference and notably lower than Test AA suggest potential overfitting and increased privacy risk, while high AA values may indicate distributional shifts. Privacy loss can be inferred as the difference between Test AA and Train AA.

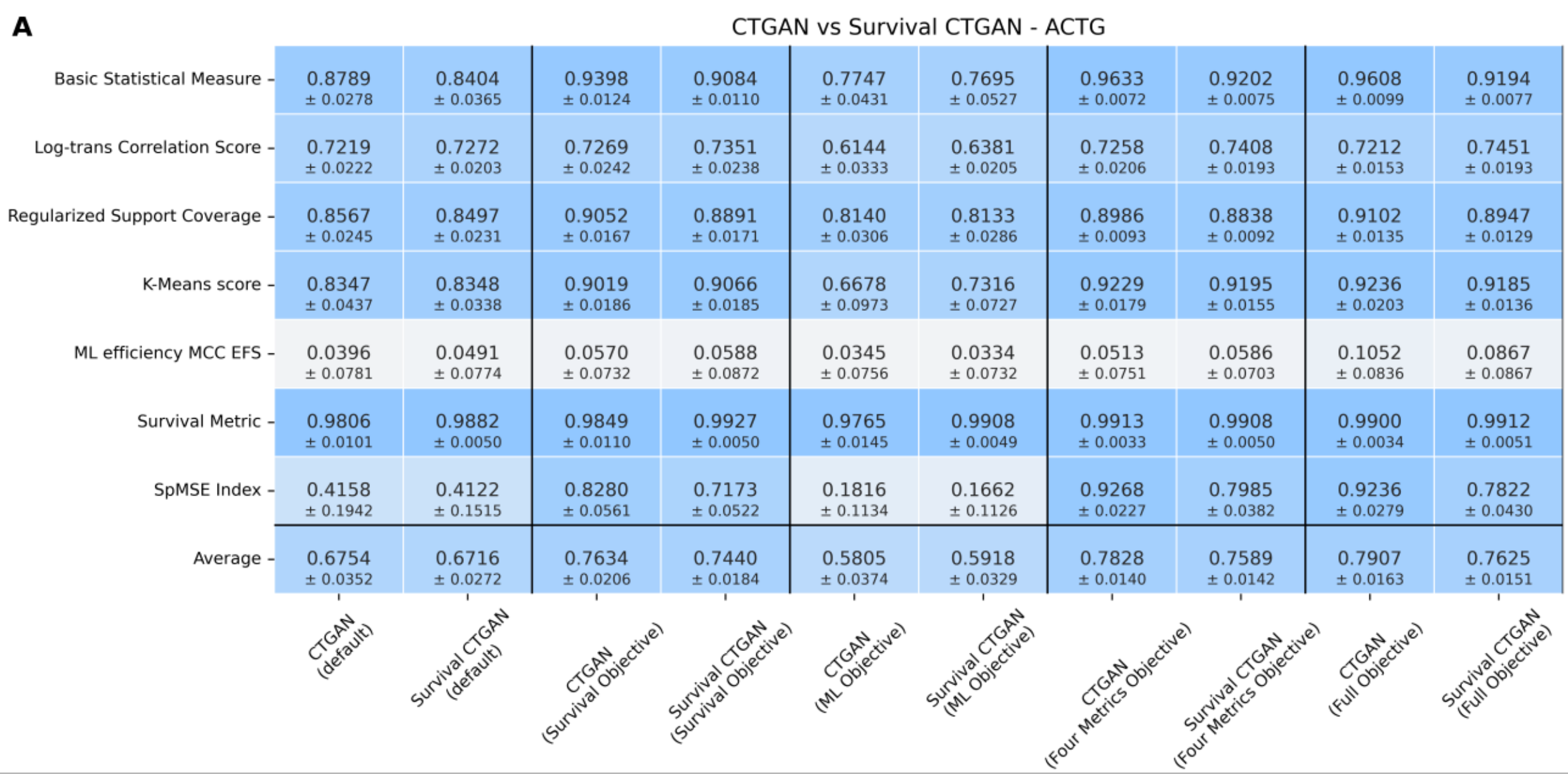

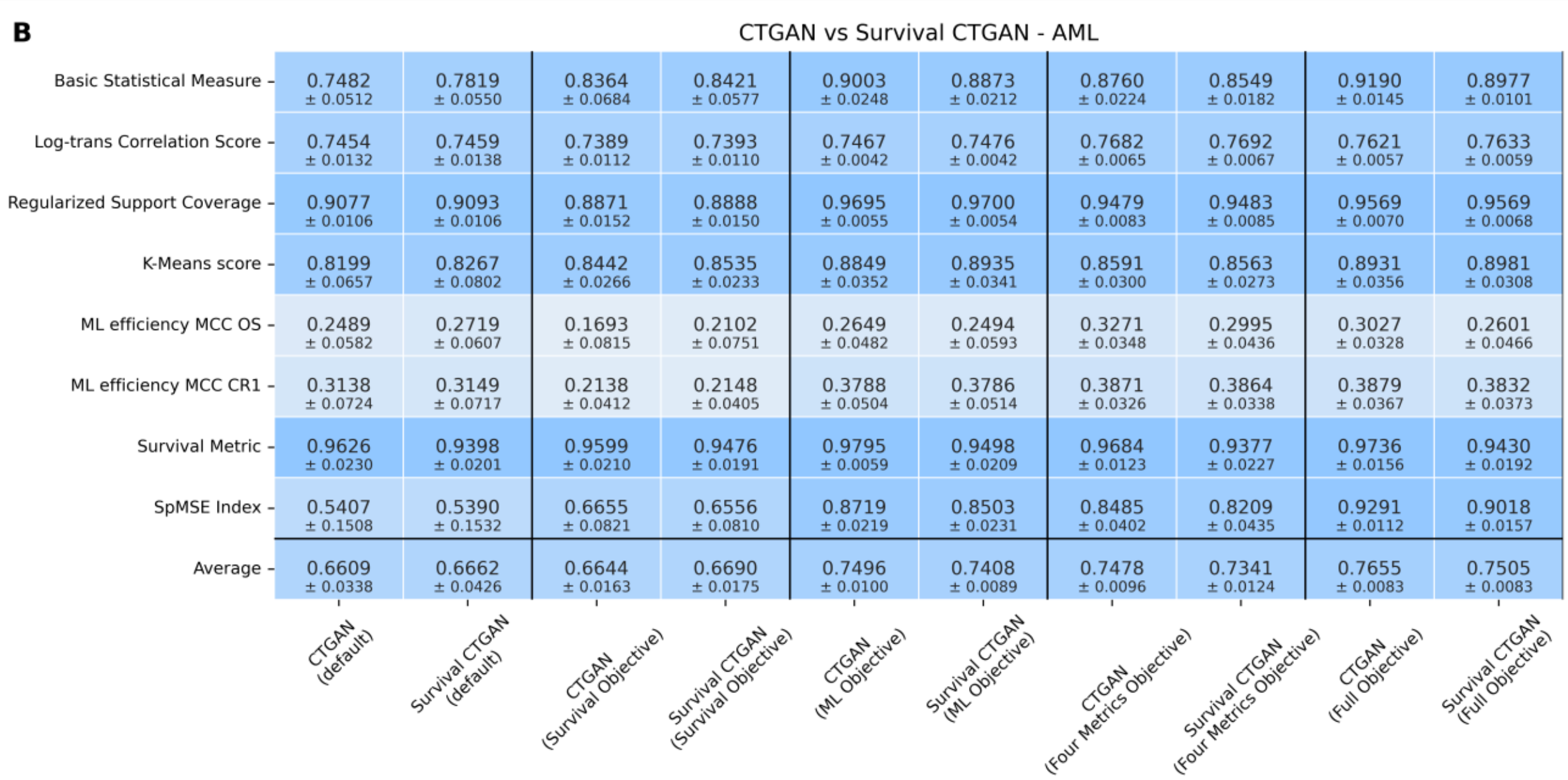

**Figure A2**. Comparison of CTGAN and Survival CTGAN performance on (A) ACTG and (B) AML datasets using identical hyperparameters, training procedures, and sampling seeds. For each metric the mean ± standard deviations of 25 synthetic datasets

are presented. While Survival CTGAN previously outperformed CTGAN in independent optimizations, this controlled comparison shows that CTGAN generally achieved higher metric scores when the same conditions were applied.

**A** Model Reevaluation after the removal of non-valid patients (best HPO) - ACTG

| | train -> test | RTVAE | TVAE | CTGAN | CTAB-GAN+ | Tab DDPM | SURVAE | Survival GAN | Survival CTGAN | Survival NFlow |
|---|---|---|---|---|---|---|---|---|---|---|
| Basic Statistical Measure | 0.9578 | 0.8783 ± 0.0064 | 0.9562 ± 0.0104 | 0.9424 ± 0.0120 | 0.9480 ± 0.0232 | 0.9281 ± 0.0259 | 0.8815 ± 0.0071 | 0.8500 ± 0.0542 | 0.9164 ± 0.0085 | 0.9065 ± 0.0116 |
| Log-trans Correlation Score | 0.6647 | 0.6149 ± 0.0212 | 0.7795 ± 0.0240 | 0.6906 ± 0.0189 | 0.7197 ± 0.0520 | 0.7497 ± 0.0221 | 0.6656 ± 0.0185 | 0.5674 ± 0.0553 | 0.7387 ± 0.0179 | 0.7384 ± 0.0471 |
| Regularized Support Coverage | 0.9038 | 0.7498 ± 0.0179 | 0.9055 ± 0.0173 | 0.8923 ± 0.0106 | 0.8900 ± 0.0205 | 0.8970 ± 0.0125 | 0.8634 ± 0.0095 | 0.7979 ± 0.0192 | 0.8631 ± 0.0143 | 0.8445 ± 0.0280 |
| K-Means score | 0.9046 | 0.7453 ± 0.0325 | 0.9172 ± 0.0181 | 0.9098 ± 0.0170 | 0.8782 ± 0.0318 | 0.9137 ± 0.0176 | 0.8761 ± 0.0210 | 0.8052 ± 0.0422 | 0.9153 ± 0.0191 | 0.9072 ± 0.0184 |
| ML efficiency MCC EFS | 0.1221 | 0.0204 ± 0.0819 | 0.0639 ± 0.0768 | 0.0583 ± 0.0902 | 0.0436 ± 0.0761 | 0.0810 ± 0.0687 | 0.0597 ± 0.1080 | 0.0189 ± 0.0670 | 0.0796 ± 0.1049 | 0.0445 ± 0.0750 |
| Survival Metric | 0.9884 | 0.9464 ± 0.0010 | 0.9897 ± 0.0042 | 0.9876 ± 0.0048 | 0.9878 ± 0.0045 | 0.9961 ± 0.0013 | 0.9898 ± 0.0054 | 0.9854 ± 0.0033 | 0.9888 ± 0.0040 | 0.9887 ± 0.0057 |
| SpMSE Index | 0.9314 | 0.7480 ± 0.0526 | 0.9162 ± 0.0401 | 0.7680 ± 0.0519 | 0.8167 ± 0.0803 | 0.9250 ± 0.0307 | 0.7953 ± 0.0554 | 0.4775 ± 0.1616 | 0.7686 ± 0.0339 | 0.7491 ± 0.0513 |
| Average | 0.7818 | 0.6719 ± 0.0157 | 0.7897 ± 0.0139 | 0.7498 ± 0.0168 | 0.7549 ± 0.0252 | 0.7844 ± 0.0117 | 0.7331 ± 0.0165 | 0.6432 ± 0.0266 | 0.7529 ± 0.0160 | 0.7398 ± 0.0167 |

**B** Model Reevaluation after the removal of non-valid patients (best HPO) - AML

| | train -> test | RTVAE | TVAE | CTGAN | CTAB-GAN+ | Tab DDPM | SURVAE | Survival GAN | Survival CTGAN | Survival NFlow |
|---|---|---|---|---|---|---|---|---|---|---|
| Basic Statistical Measure | 0.9454 | 0.7234 ± 0.1448 | 0.9083 ± 0.0234 | 0.6572 ± 0.0374 | 0.8133 ± 0.1279 | 0.9003 ± 0.0238 | 0.8408 ± 0.0404 | 0.8779 ± 0.0267 | 0.8956 ± 0.0108 | 0.8675 ± 0.0215 |
| Log-trans Correlation Score | 0.5773 | 0.6110 ± 0.2626 | 0.7260 ± 0.0451 | 0.6699 ± 0.0351 | 0.7646 ± 0.0236 | 0.6978 ± 0.0223 | 0.7363 ± 0.0362 | 0.7488 ± 0.0112 | 0.7665 ± 0.0061 | 0.7291 ± 0.0190 |
| Regularized Support Coverage | 0.9344 | 0.8044 ± 0.1385 | 0.8863 ± 0.0161 | 0.9276 ± 0.0128 | 0.9383 ± 0.0104 | 0.8622 ± 0.0072 | 0.8540 ± 0.0202 | 0.9426 ± 0.0097 | 0.9532 ± 0.0070 | 0.9627 ± 0.0090 |
| K-Means score | 0.8859 | 0.7099 ± 0.0879 | 0.9038 ± 0.0233 | 0.8269 ± 0.0347 | 0.8812 ± 0.0802 | 0.9310 ± 0.0119 | 0.8230 ± 0.0748 | 0.8697 ± 0.0297 | 0.8955 ± 0.0304 | 0.8741 ± 0.0196 |
| ML efficiency MCC OS | 0.3079 | 0.2309 ± 0.1329 | 0.2930 ± 0.0675 | 0.2529 ± 0.0585 | 0.2941 ± 0.0636 | 0.3269 ± 0.0401 | 0.2181 ± 0.0601 | 0.2353 ± 0.0479 | 0.2551 ± 0.0475 | 0.2026 ± 0.0608 |
| ML efficiency MCC CR1 | 0.4223 | 0.2495 ± 0.1461 | 0.3138 ± 0.0448 | 0.2412 ± 0.0559 | 0.3493 ± 0.0781 | 0.3968 ± 0.0418 | 0.3299 ± 0.0502 | 0.3597 ± 0.0502 | 0.3800 ± 0.0353 | 0.3281 ± 0.0411 |
| Survival Metric | 0.9769 | 0.7941 ± 0.0798 | 0.9666 ± 0.0166 | 0.8727 ± 0.0400 | 0.9587 ± 0.0359 | 0.9913 ± 0.0040 | 0.9137 ± 0.0474 | 0.9552 ± 0.0197 | 0.9502 ± 0.0181 | 0.9344 ± 0.0248 |
| SpMSE Index | 0.9158 | 0.7886 ± 0.0691 | 0.9224 ± 0.0185 | 0.8243 ± 0.0259 | 0.8792 ± 0.0492 | 0.9286 ± 0.0164 | 0.8185 ± 0.0697 | 0.8621 ± 0.0226 | 0.9074 ± 0.0125 | 0.8245 ± 0.0333 |
| Average | 0.7457 | 0.6140 ± 0.1237 | 0.7400 ± 0.0134 | 0.6591 ± 0.0196 | 0.7349 ± 0.0502 | 0.7544 ± 0.0107 | 0.6918 ± 0.0393 | 0.7314 ± 0.0127 | 0.7505 ± 0.0102 | 0.7154 ± 0.0090 |

**Figure A3.** Reevaluation of generative model performance on (A) ACTG and (B) AML datasets after removing non-valid patients (*V7*). The heatmaps show the mean ± standard deviations of metrics for each model using their best HPO strategy. While most models experienced slight decreases in performance, the $S_{pMSE}$ Index improved, indicating reduced distinguishability from real data. The impact of patient removal was more pronounced for general-purpose models, particularly on the AML dataset, highlighting the role of preprocessing in ensuring logical consistency. Color intensity reflects deviation from the real-data reference: values closer to the real baseline are shown in neutral tones, higher values are shaded blue, and lower values are shaded red.

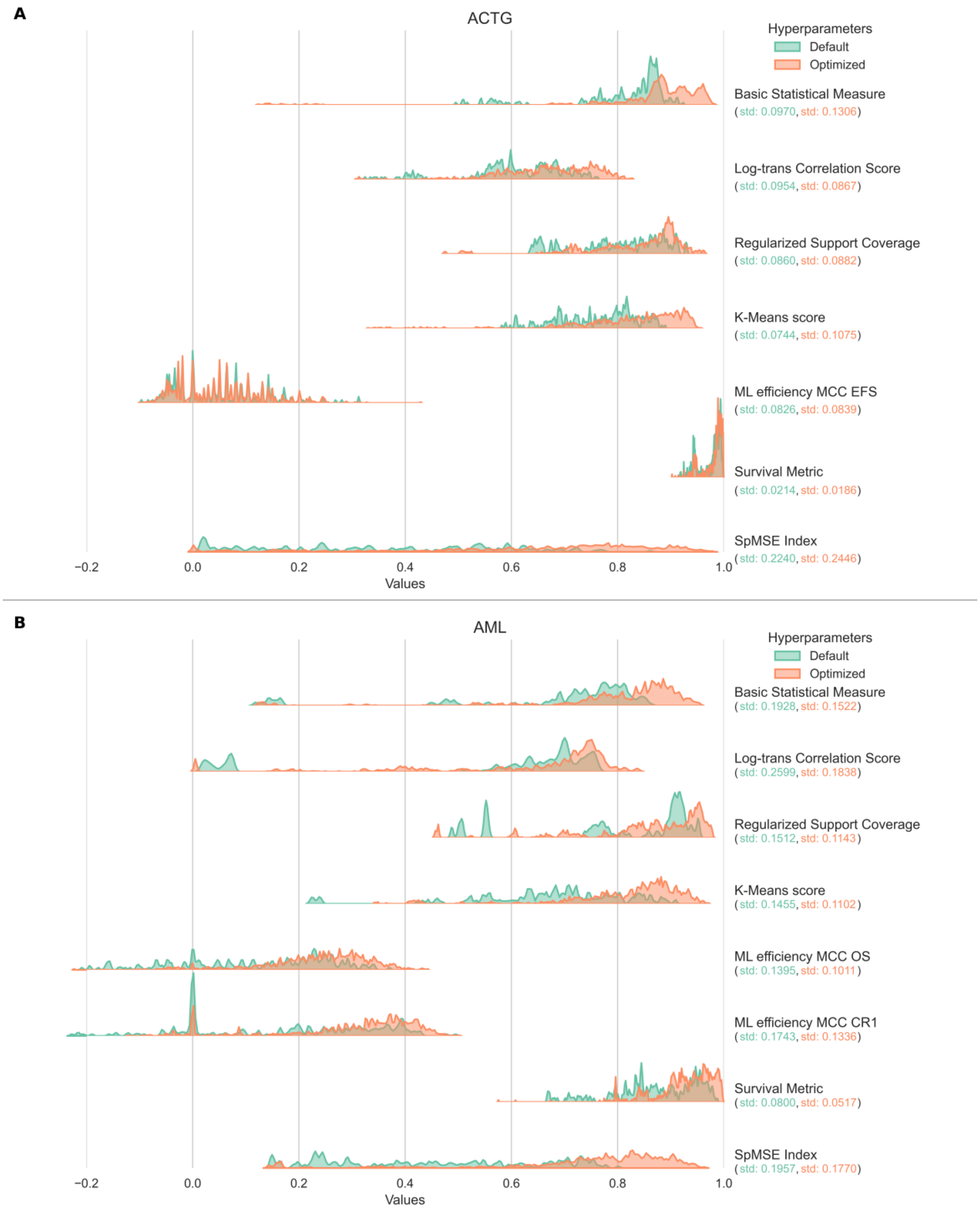

**Figure A4**. Distribution of data quality evaluation metrics for default and optimized hyperparameter configurations across synthetic datasets for (A) ACTG and (B) AML. Default represents all synthetic datasets produced by the nine generative models (225) and optimized all the synthetic datasets produced by models with one of the four HPO objectives, resulting in 900 synthetic datasets.

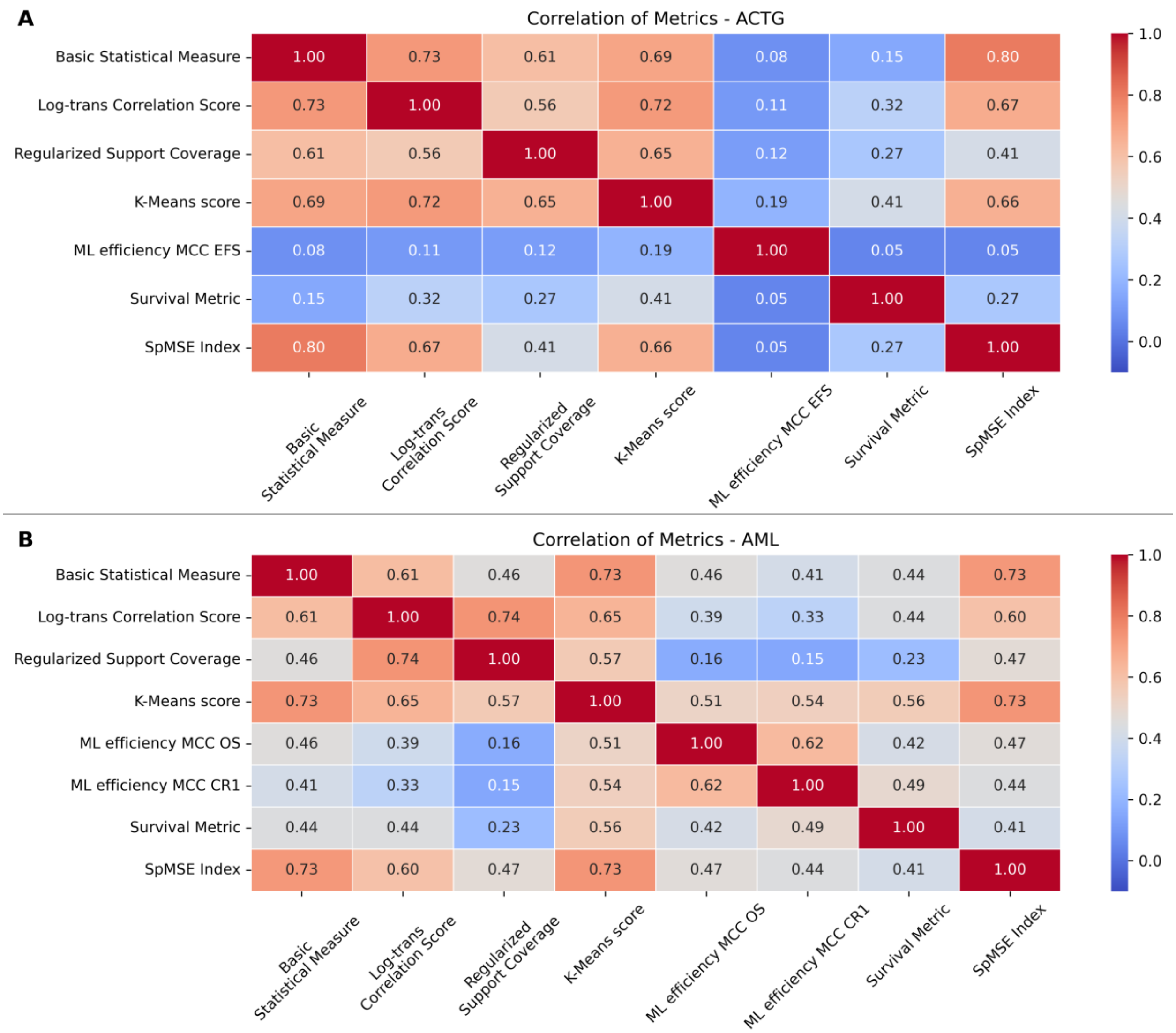

**Figure A5**. Correlation matrices of evaluation metrics for (A) ACTG and (B) AML datasets. Strong correlations between metrics such as the Basic Statistical Measure, K-Means Score, and Log-transformed Correlation Score suggest potential redundancy in compound metric optimization targets. In contrast, metrics like the Survival Metric and ML Efficiency scores exhibit weaker correlations, indicating they capture more independent characteristics.